\documentclass[runningheads]{llncs}

\usepackage{eccv}

\usepackage{eccvabbrv}

\usepackage{graphicx}
\usepackage{booktabs}       
\usepackage{amsfonts}       
\usepackage{nicefrac}       
\usepackage{microtype}      
\usepackage{xcolor}         
\usepackage{colortbl}
\usepackage{amsmath}
\usepackage{tikz}
\usepackage{comment}
\usepackage{color}

\usepackage{enumitem}
\usepackage{graphicx}
\usepackage{amsmath}
\usepackage{booktabs}
\usepackage{mathtools}
\usepackage{url}
\usepackage{comment}
\usepackage{color}
\usepackage{graphicx,setspace}
\usepackage{multirow, array, booktabs,makecell,xspace,bm}
\usepackage{pifont}
\usepackage{amsfonts}
\usepackage{arydshln}
\usepackage{dsfont}
\usepackage{stackengine}
\usepackage{scalerel}
\usepackage{esdiff} 
\usepackage{bbm}

\newcommand{\cmark}{\ding{51}}%
\newcommand{\xmark}{\ding{55}}%

\usepackage[accsupp]{axessibility}  

\usepackage[table]{xcolor}

\usepackage{tikz}
\usepackage{comment}
\usepackage{color}

\usepackage{enumitem}
\usepackage{graphicx}
\usepackage{amsmath}
\usepackage{booktabs}
\usepackage{mathtools}
\usepackage{url}
\usepackage{comment}
\usepackage{color}
\usepackage{graphicx,setspace}
\usepackage{multirow, array, booktabs,makecell,xspace,bm}
\usepackage{pifont}
\usepackage{amsfonts}
\usepackage{arydshln}
\usepackage{dsfont}
\usepackage{xcolor}
\usepackage{colortbl}

\usepackage{stackengine}

\usepackage{scalerel}
\usepackage{esdiff} 

\usepackage{bbm}
\usepackage{makecell}

\usepackage[most]{tcolorbox} 
\usepackage{tcolorbox}
\usepackage[x11names]{xcolor}
\usepackage{soul}
\sethlcolor{pink}

\newtcolorbox{mybox}[1][]{%
    enhanced,
    top=1pt,
    bottom=1pt,
    boxrule=0pt,
    arc=0pt,
    colframe=white,
    colback=white,
    borderline north={0.5mm}{0mm}{black},
    borderline south={0.5mm}{0mm}{black},
    #1
}
\newtcolorbox{mybox2}[1][]{%
    enhanced,
    top=1pt,
    bottom=1pt,
    boxrule=0pt,
    arc=0pt,
    colframe=white,
    colback=white,
    borderline north={0.5mm}{0mm}{black},
    borderline south={0.5mm}{0mm}{black},
    width=\textwidth,
    left=0pt,
    right=0pt,
    #1
}

\renewcommand{\thetable}{S.\arabic{table}}

\renewcommand{\thefigure}{S.\arabic{figure}}

\appendix

\usepackage{hyperref}

\makeatletter
\newcommand{\blfootnote}[1]{%
  \begingroup
    \renewcommand\thefootnote{}%
    \renewcommand\@makefnmark{}%
    \footnotetext{#1}%
  \endgroup
}
\makeatother

\usepackage{orcidlink}

\newcommand{\equalcontrib}{\textsuperscript{*}}
\newcommand{\corrauthor}{\textsuperscript{$\dagger$}}

\begin{document}

\title{Different Changes Require Different Reasoning:
Change-Type-Specialized Experts for Robust Change Captioning} 

\titlerunning{MEDIC}

\author{
Jiyoung Park\equalcontrib\orcidlink{0009-0000-4212-7745}
\and
InJae Oh\equalcontrib\orcidlink{0009-0001-1321-3019}
\and
Jung Uk Kim\corrauthor\orcidlink{0000-0003-4533-4875}
}

\authorrunning{J. Park et al.}

\institute{
Kyung Hee University, Yong-in, South Korea \\
\email{\{jy0117, seanoh, ju.kim\}@khu.ac.kr}
}

\maketitle

\blfootnote{\textsuperscript{*} Equal contribution. \quad
\textsuperscript{$\dagger$} Corresponding author.}

\begin{abstract}
Change captioning is the task of generating natural language descriptions that explain the changes between a pair of images. Although different change types (\textit{e.g.,} color shifts, object additions) exhibit distinct visual cues and require specialized reasoning processes, existing methods often overlook these distinctions. To address this limitation, we propose Multi-Expert Diagnosis for Image Change (MEDIC), a novel framework that introduces change-type awareness by explicitly modeling change categories. We build our MEDIC as a memory network to dynamically retrieve type-relevant visual patterns conditioned on the input. This design allows each expert to flexibly capture diverse variations within each change type and focus on the most informative cues for its designated change type. By routing inputs through type-specialized experts and learning dedicated representations for each change category, MEDIC generates more precise and type-aware change descriptions. Extensive experiments demonstrate that proposed MEDIC consistently outperforms across diverse and challenging datasets. The code is
available at https://github.com/VisualAIKHU/MEDIC.
\end{abstract}

\section{Introduction}
\label{sec:intro}

In many vision-based applications, understanding how a scene evolves over time is essential. 
Change captioning~\cite{DUDA, spotthediff} addresses this need by generating natural language descriptions that explain the visual differences between two images taken at different times, enabling intuitive communication of what has changed and how it appears. 
Such capability is particularly valuable in real-world scenarios such as surveillance~\cite{surveillance2,spotthediff} and medical diagnostics~\cite{medicalRef}, where precise and reliable interpretation of visual changes is critical.

Research in change captioning has progressed through several stages: early pixel- or feature-level differencing with simple encoder-decoder captioning~\cite{DUDA, spotthediff}, improved alignment and robustness via cross-view matching, distractor suppression, and structure-aware modeling~\cite{dirl, smart, scorer}, and more recent generative and reasoning-centric approaches that incorporate contrastive or diffusion objectives, adversarial hard negatives, MLLM/VLM-based reasoning, and textual compositional reasoning~\cite{decider, rdd, vixen, blip2idc, cortex}.

\begin{figure}[t]
\centering

\begin{minipage}[t]{0.64\linewidth}
\vspace{0pt}
\centering
\includegraphics[width=\linewidth]{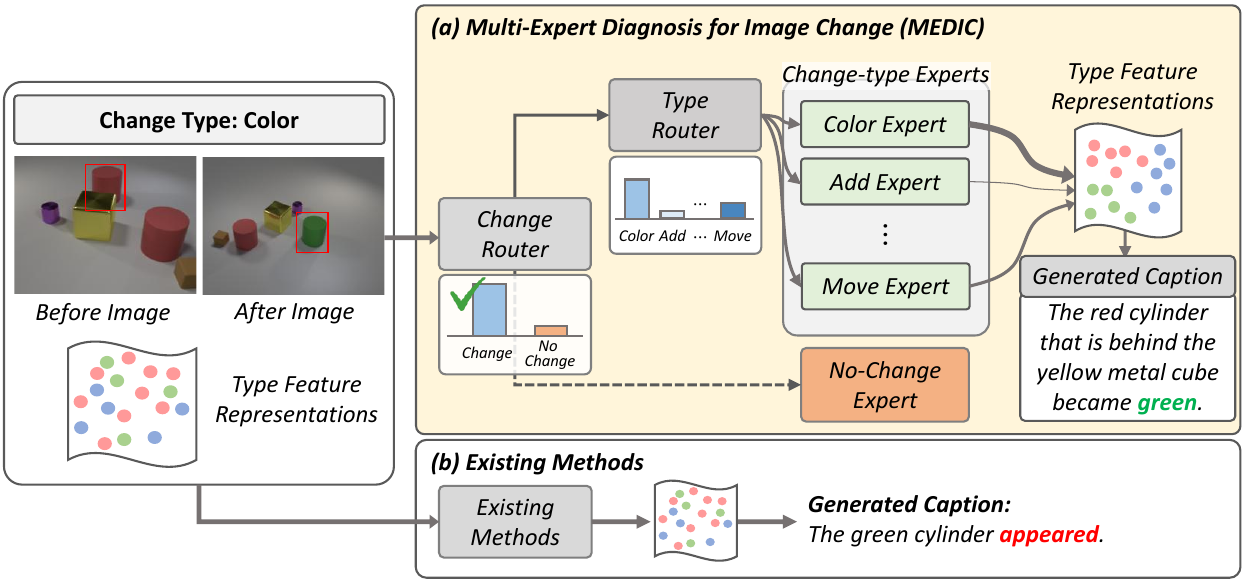}
\captionsetup{skip=3pt}
\captionof{figure}{Conceptual comparison between (a) MEDIC and (b) existing methods using a color-change example.}
\label{fig:concept}
\end{minipage}
\hfill
\begin{minipage}[t]{0.346\linewidth}
\vspace{-7.5pt}
\centering
\includegraphics[width=\linewidth]{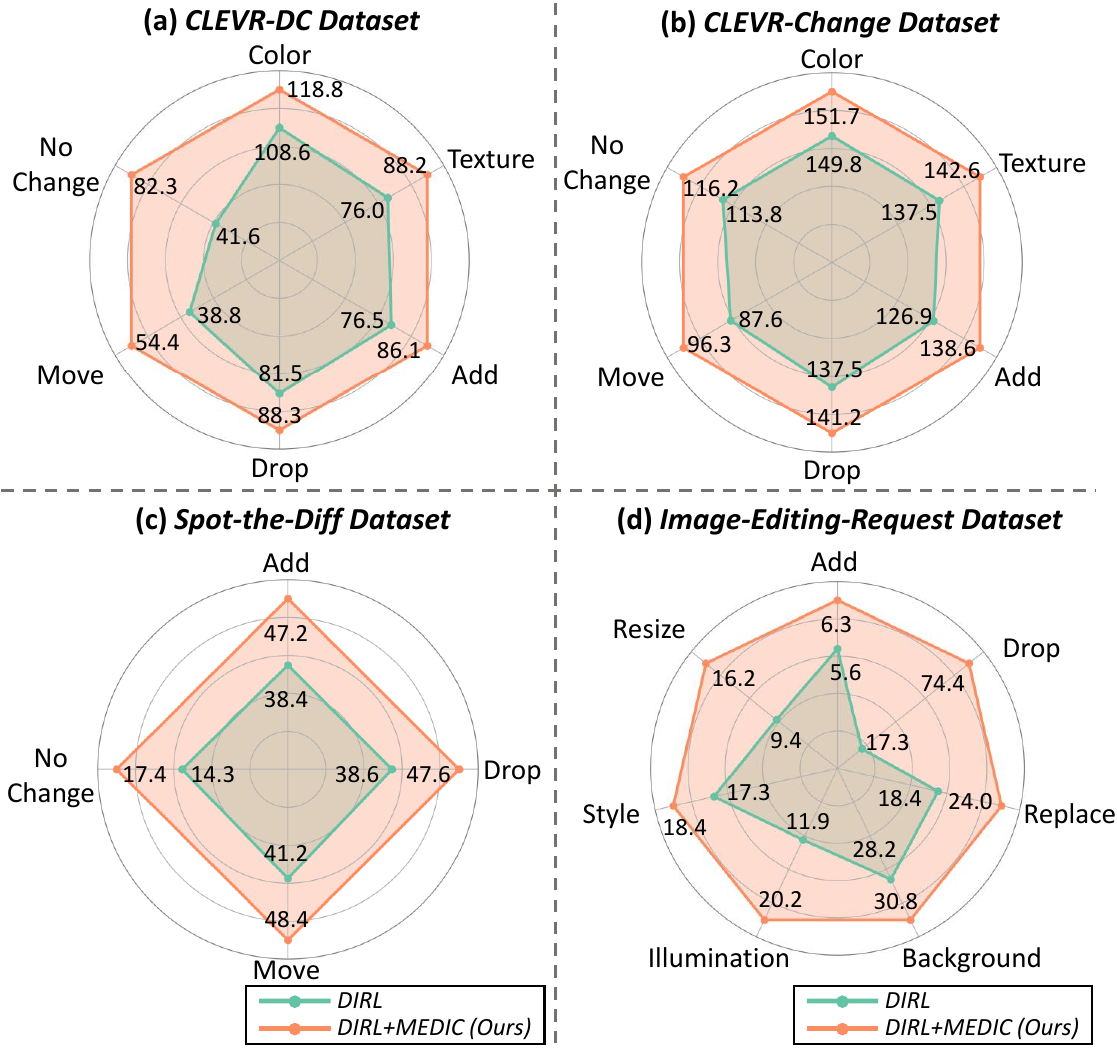}
\captionsetup{skip=3pt}
\captionof{figure}{CIDEr comparisons across change types.}
\label{fig:radar}
\end{minipage}

\vspace{-0.4cm}
\end{figure}

Despite these advances, most prior methods still assume that diverse visual changes can be addressed through a single reasoning process. Color change depends on subtle pixel-level intensity variations~\cite{color_detect_cue}, and object addition requires background separation and object-level reasoning~\cite{add_detect_cue}. Nevertheless, existing methods are constrained to process color variations, object additions, and spatial movements through a single pipeline even though each type relies on different visual evidence.
As shown in Fig.~\ref{fig:concept} (b), when these heterogeneous cues are processed within a single unified pipeline, the model often misinterprets the nature of the change-type (\textit{e.g.,} describing a red cylinder that became green (color change-type) as `the green cylinder appeared' (add change-type) rather than `the red cylinder that is behind the yellow metal cube became green'). 
They overlook the fact that different change types rely on distinct visual cues. 

To address this issue, we introduce \textbf{MEDIC} (\textbf{M}ulti-\textbf{E}xpert \textbf{D}iagnosis for \textbf{I}mage \textbf{C}hange), a new framework that models change captioning as a type-aware process. 
As shown in Fig.~\ref{fig:concept} (a), MEDIC employs a two-stage routing mechanism inspired by mixture-of-experts (MoE)~\cite{MoE}: a change router first identifies whether a meaningful change is present, and a type router then assigns the input to a specialized expert dedicated to the corresponding change type. 
We design each expert as a memory network that retrieves diverse type-relevant visual patterns, enabling adaptive and discriminative reasoning beyond conventional feed-forward experts~\cite{MoE}. 
To support type-aware specialization, we introduce three training losses—routing loss, expert consistency loss, and expert disentangle loss—that collectively promote correct expert assignment, enforce type-specific representation learning, and maintain clear separation among expert representations. 
As a result, MEDIC consistently outperforms existing state-of-the-art, code-released methods such as DIRL~\cite{dirl} across all change types and datasets (see Fig.~\ref{fig:radar}), showing the effectiveness of explicitly modeling change types in change captioning.

The main contributions of our work are as follows:
\begin{itemize}
    \item We introduce \textbf{MEDIC}, a novel type-aware change captioning framework that addresses the limitations that does not consider change types. By using specialized experts for each change category, MEDIC prevents reasoning confusion and enables type-aware interpretation.
    \item We design specialized memory-based experts that dynamically retrieve input-dependent patterns within each change type, resulting in consistent performance gains across all datasets and change types.
    \item We develop three training losses—routing loss, expert consistency loss, and expert disentangle loss—to learn type-specific representations while encouraging clear separation across change categories.
\end{itemize}

\section{Related Works}
\label{sec:RelatedWorks}

\subsection{Change Captioning}
Change captioning generates descriptions of semantic differences between image pairs. Early methods used aligned pairs~\cite{spotthediff} or feature subtraction~\cite{DUDA}, but struggled with distractors such as viewpoint shifts.
Recent methods improve robustness through relation-aware difference modeling and distractor suppression: SRDRL~\cite{SRDRL} and R$^3$Net~\cite{r3net} learn semantic or relation-embedded differences, SCORER~\cite{scorer} and DIRL~\cite{dirl} use contrastive and relational learning, SMART~\cite{smart}, NCT~\cite{NCT}, and I3N~\cite{i3n} exploit syntactic, neighborhood, or cross-view cues, and MURAT~\cite{MURAT} and CHEERS~\cite{cheers} further improve multi-grained aggregation and change-entity-guided disentanglement.

Despite these advances, most methods still use a unified, type-agnostic pipeline. They do not explicitly adapt visual difference reasoning to the change type MEDIC addresses this limitation with type-specific experts.

\subsection{Mixture of Experts}
The Mixture of Experts (MoE) framework routes inputs to specialized sub-networks, or `experts,' each handling a subset of the input space~\cite{jacobs1991adaptive}. Sparsely-Gated MoE~\cite{shazeer2017outrageously} enabled scalable training by activating only a few experts per input. This design was later adopted in large-scale models such as GShard~\cite{lepikhin2020gshard} and Switch Transformers~\cite{fedus2022switch} to boost capacity without proportional compute.

Beyond language modeling, MoE has been effective in vision/multimodal learning. V-MoE~\cite{riquelme2021scaling} introduced sparsely activated experts into Vision Transformers, and later works extended MoE to structured domains like scene decomposition~\cite{zhenxing2022switch} and visual tracking~\cite{cai2025spmtrack}, validating its strength in handling diverse inputs and enabling specialization.
Unlike conventional MoE designs that rely on shared representations between experts, MEDIC specializes in distinct change categories.

\subsection{Memory Network}

Memory networks were introduced to equip models that support content-based retrieval. Notable examples include Dynamic Memory Networks \cite{kumar2016ask} and Key-Value Memory Networks \cite{miller2016key}, which improve reasoning by explicitly modeling memory access mechanisms. In computer vision, such networks have been applied to various tasks, including object tracking \cite{zhu2020inflated}, predictive representation learning \cite{han2020memory}, and video question answering \cite{fan2019heterogeneous}. Key-value memory, in particular, enables efficient and interpretable retrieval through query-key matching \cite{marchetti2020mantra, lee2021video}, and recent works have extended this design by incorporating multi-modal cues or spatial priors for fine-grained alignment \cite{kim2023enabling, weng2025pattern, kim2022towards, kim2021robust, kim2023stereoscopic}.

Motivated by these advances, we design each expert as a key-value memory. It enables type-specific reasoning by dynamically attending to relevant features, supporting more adaptive and semantically grounded change reasoning.

\section{Methodology}
\label{sec:Method}

\begin{figure*}[t]
    \begin{minipage}[b]{\linewidth}
	\centering
        \centerline{\includegraphics[width=\linewidth]{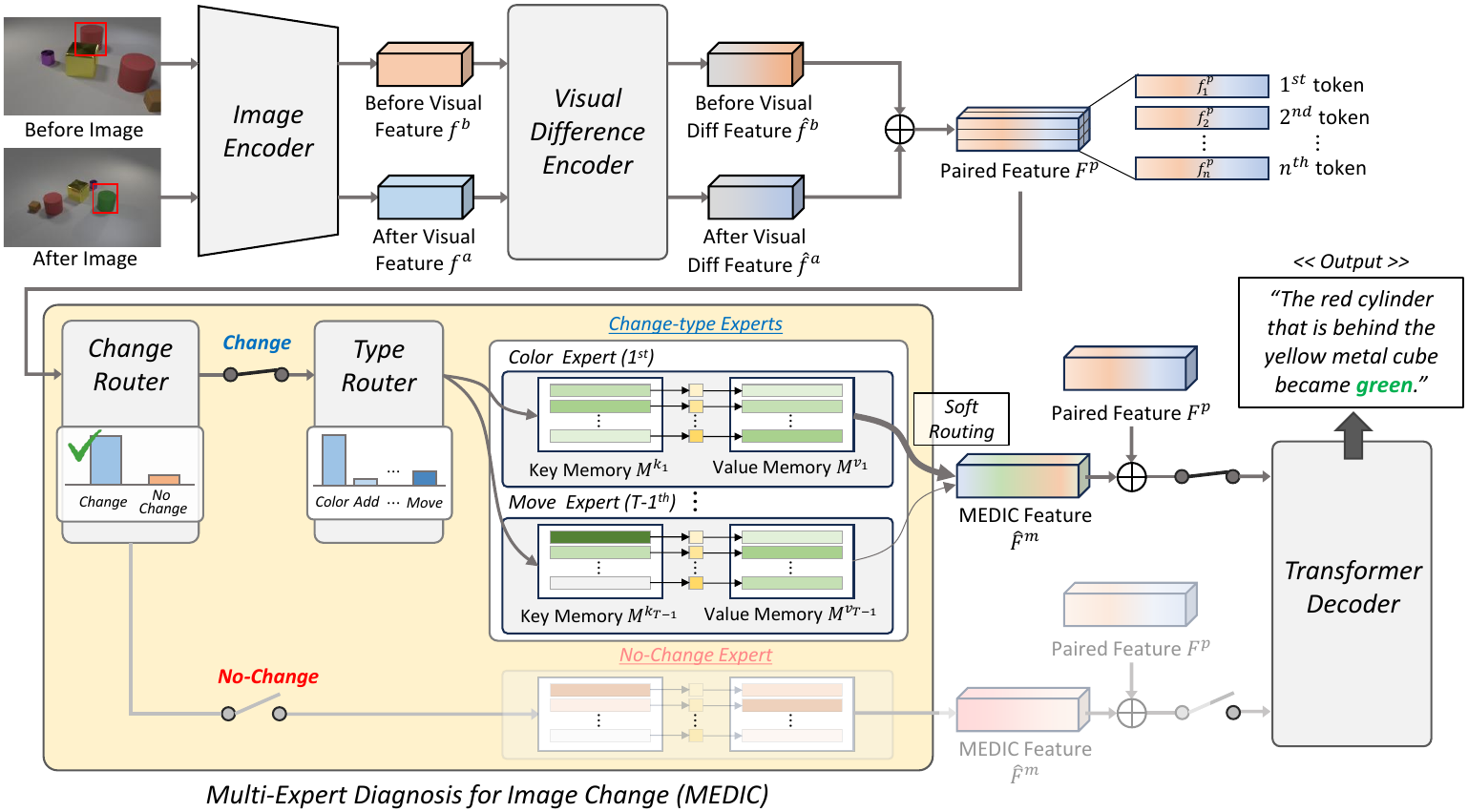}}
        \end{minipage}
    \centering
    \vspace{-0.4cm}
	\caption{Overview of our Multi-Expert Diagnosis for Image Change (MEDIC) with a color change example. MEDIC consists of (1) a two-stage router, which first determines the presence of change and then assigns input to specialized experts based on the predicted change type, and (2) change-type experts, each designed to capture fine-grained, type-specific visual cues via dynamic memory-based retrieval. $\bigoplus$ denotes concatenation.}
    \vspace{-0.2cm}
    \label{fig:main_architecture}
\end{figure*}

\subsection{Overall Architecture}
As shown in Fig. \ref{fig:main_architecture}, we propose MEDIC to generate accurate and type-aware change descriptions by explicitly modeling change-type diversity.
We describe how MEDIC can be integrated into existing change captioning pipelines that use conventional visual difference encoders and transformer decoders.
Given a pair of before image $I^b$ and after image $I^a$, visual features $f^b$ and $f^a$ are first extracted using the same image encoder, ResNet-101 \cite{ResNet101}. These features are passed to a visual difference encoder that highlights potential change cues between the two images, referred to as visual difference features $\hat{f}^b$ and $\hat{f}^a$. They are concatenated to form paired feature $F^p {=} [\hat{f}^b;\hat{f}^a] \in \mathbb{R}^{h \times w \times 2d}$ ($h$, $w$, and $d$ indicate height, width, and feature dimension, respectively). Then it is fed into the MEDIC.

In the MEDIC, two-stage routers guide the interaction with a set of $T$ change-type experts. A change router first detects whether a change has occurred. If the change is detected, then a type router estimates the distribution over the ($T{-}1$) actual change types (\textit{e.g.,} color, texture, add) and softly routes the paired feature $F^p$ to the change-type experts based on the estimated type distribution. The set of expert outputs are denoted as $F^m {=} \{F^{m_t}\}_{t=1}^{T{-}1}$ where each $F^{m_t} \in \mathbb{R}^{h\times w\times 2d}$. These outputs are then aggregated via a weighted summation according to the predicted type distribution to form a MEDIC feature $\hat{F}^m \in \mathbb{R}^{h\times w\times 2d}$. In contrast, if no-change is detected, only the output of the no-change expert ($T$-th expert) is used as $\hat{F}^m$. Then, $\hat{F}^m$ and $F^p$ are concatenated and fed into a transformer decoder to generate the change caption.

\subsection{Two-stage Router}
The two-stage router in MEDIC routes the paired feature $F^p$ based on a two-step decision process: (1) the change router determines whether a change is present, and (2) if so, the type router estimates the distribution over change types. \\

\noindent\textbf{(1) Change Router.}
To determine whether a change is present, the change router applies a classifier to the input $F^p$. For the $i$-th sample, it encodes pixel-level logits $\mathbf{z}_\text{chg}^i \in \mathbb{R}^{h \times w \times 2}$, then aggregated via global average pooling (GAP) and passed through a softmax to obtain the change probability vector $p^i \in \mathbb{R}^2$.

The change router is trained using cross-entropy loss: 
\begin{equation}
    \mathcal{L}_{chg} = - \frac{1}{B} \sum_{i=1}^{B} \log p^i_{y^i},
\end{equation}
where $B$ is the batch size, $y^i \in \{0,1\}$ denotes the ground-truth label for the $i$-th input (1: change, 0: no-change), and $p^i_{y^i}$ is the predicted probability in $p^i$. \\

\noindent\textbf{(2) Type Router.}
For inputs where a change is detected, the type router takes $F^p$ as input and outputs pixel-level logits $\mathbf{z}_\text{type}\in\mathbb{R}^{h\times w\times(T-1)}$. These logits are pooled with GAP and passed through softmax to yield the change-type probability vector $p^i_{\text{type}} \in \mathbb{R}^{T-1}$. Here, $(T{-}1)$ denotes the number of change types, excluding the no-change category.
The type router is also trained using the cross-entropy loss:
\begin{equation}
    \mathcal{L}_{type} = - \frac{1}{B_{\text{chg}}} \sum_{i=1}^{B_{\text{chg}}} \log p^i_{\text{type}, y^i},
\end{equation}
where $B_{\text{chg}}$ is the number of samples with actual changes, $y^i \in \{1,2,\dots,T{-}1\}$ is the ground-truth change-type label for the $i$-th sample, and $p^i_{\text{type}, y^i}$ is the predicted probability for the correct change type.

The final routing loss $\mathcal{L}_{router}$ is defined as:
\begin{equation}
    \mathcal{L}_{router} = \mathcal{L}_{chg} + \mathcal{L}_{type},
\end{equation}
The $\mathcal{L}_{router}$ encourages the router to adaptively select and apply suitable reasoning according to the change type.

\subsection{Memory Network for Change-type Experts}
\begin{figure}[t]
    \begin{minipage}[b]{\linewidth}
	\centering
        \centerline{\includegraphics[width=0.7\linewidth]{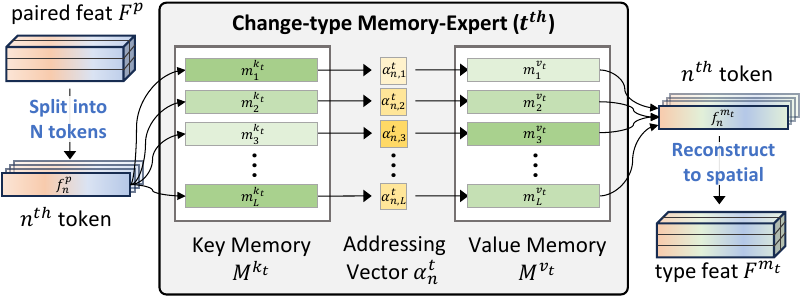}}
        \end{minipage}
    \vspace{-0.5cm}
    \caption{Detailed architecture of the change-type memory-expert. Paired feature $F^p$ is split into $N$ tokens, each independently processed through a memory network (illustrated here with the $n$-th token). Each token retrieves type-specific cues via addressing, and the outputs are reassembled into the spatial feature $ F^{m_t}$.}
    \label{fig:memory_network}
    \vspace{-0.5cm}
\end{figure}

We design expert, specialized for a specific change type $t\in\{1,...,T\}$, as a key-value memory network to capture distinct visual cues. As shown in Fig. \ref{fig:memory_network}, the memory of the $t$-th expert is defined as $M^t=\{M^{k_t}, M^{v_t}\}$, where $M^{k_t}=\{m_{\ell}^{k_t}\}_{\ell=1}^L$ and $M^{v_t} = \{m_{\ell}^{v_t}\}_{\ell=1}^L$ denote the key and value memory, each with $L$ slots of dimension $\mathbb{R}^{2d}$.

To embed diverse grid-level representations of each change type, we divide the paired feature $F^p$ into $N {=} hw$ tokens $\{f_n^p\}_{n=1}^N$  where each $f_n^p \in \mathbb{R}^{2d}$. For each token, we compute cosine similarity with every key memory slot to identify relevant entries for retrieval, denoted as:
\begin{equation}
    d(f_n^p, m^{k_t}_\ell) = \frac{f_n^p \cdot m_{\ell}^{k_t}}{\|f_n^p\| \|m_{\ell}^{k_t}\|}, \quad \text{for } \ell = 1,...,L.
\end{equation}

To retrieve information from the value memory, we convert these similarity scores into normalized read weights over memory slots, which we refer to as the addressing vector $\alpha_{n}^{t} {=} \{\alpha_{n,\ell}^{t}\}_{\ell=1}^L \in \mathbb{R}^{L}$. Specifically, it is computed by a softmax with temperature $\tau$:
\begin{equation}
\alpha_{n,\ell}^{t} = \frac{\exp \left( d(f_n^p, m_{\ell}^{k_t}) / \tau \right)}{\sum_{s=1}^{L} \exp \left( d(f_n^p, m_{s}^{k_t}) / \tau \right)}.
\end{equation}
Using this addressing vector, the output token $f_n^{m_t}$ is retrieved by a weighted sum over the value memory:
\begin{equation}
    f_n^{m_t} = \sum_{\ell=1}^{L} \alpha_{n,\ell}^{t} \cdot m_{\ell}^{v_t}.
\end{equation}

Applying this retrieval process to all $N$ tokens yields output tokens $\{f_n^{m_t}\}_{n=1}^N$, which are reshaped to the original spatial layout to obtain the type feature $F^{m_t} \in \mathbb{R}^{h\times w\times 2d}$ for the $t$-th expert. The memory-based design allows each expert to dynamically retrieve input-adaptive patterns for its change type, enhancing generalization across diverse scenes.

Finally, the ouput of the MEDIC feature $\hat{F}^m\in\mathbb{R}^{h\times w\times 2d}$ is obtained by aggregating the expert outputs based on the routing decisions made by the change and type routers. The aggregation is defined as follows:
\begin{equation}
    \hat{F}^{m} =
    \begin{cases}
        \sum_{t=1}^{T{-}1} p_{\text{type}, t} \cdot F^{m_t}, & \text{if a change is detected}, \\
        F^{m_T}, & \text{otherwise},
    \end{cases}
\end{equation}
where $F^{m_T}$ denotes the no-change expert output. This soft aggregation adaptively integrates expert outputs based on the predicted change type distribution.

\begin{figure}[t]
    \begin{minipage}[b]{1.0\linewidth}
	\centering
        \centerline{\includegraphics[width=\linewidth]{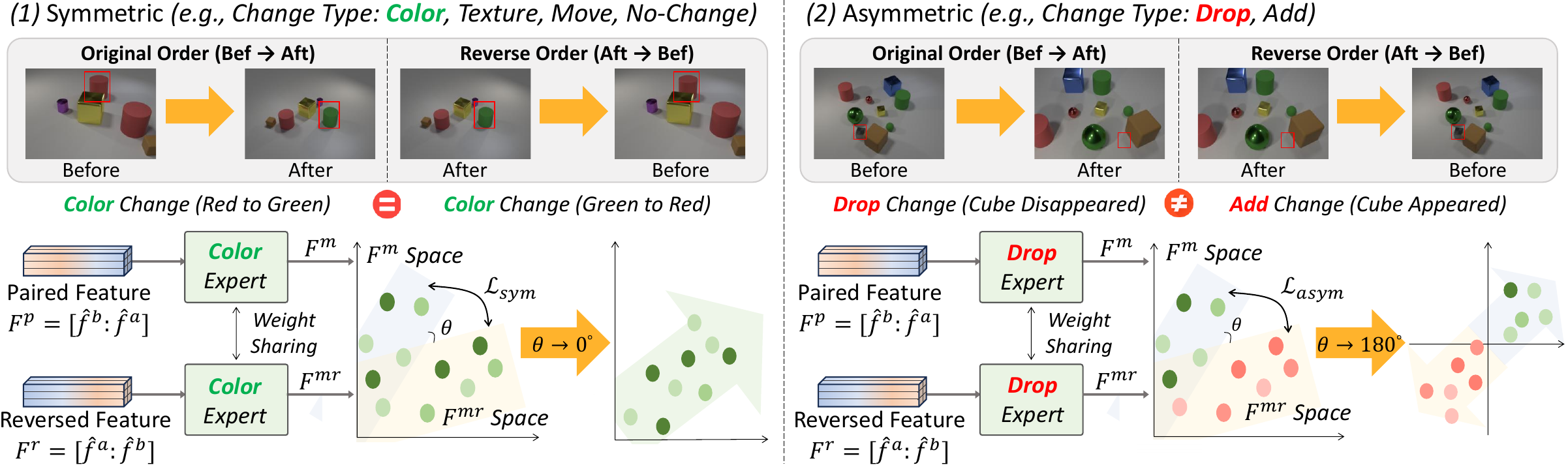}}
        \end{minipage}
    \vspace{-0.4cm}
    \caption{Overview of the expert consistency loss for (1) symmetric and (2) asymmetric change types, guiding each expert to capture its change-type behavior. For each input pair, paired and reversed features are passed through the expert of the ground-truth change type. Symmetric types (\textit{e.g.,} `color') generate type-invariant representations under reversal, whereas asymmetric types (\textit{e.g.,} `drop') yield type-variant representations.}
    \label{fig:loss}
    \vspace{-0.35cm}
\end{figure}

\subsection{Expert Consistency Loss}
To encourage each change-type expert to learn more type-specific representations, we introduce an expert consistency loss that exploits the intrinsic structural property of change types under input reversal.

As shown in Fig.~\ref{fig:loss} (1), a `color' change from red to green remains a `color' change when reversed to green to red. Since the type label is preserved under swapping, we define such types as \textit{symmetric}.
In contrast, as shown in Fig.~\ref{fig:loss} (2), a `drop' becomes an `add' when the input order is reversed. Because the type label changes under reversal, we define such cases as \textit{asymmetric}.

Based on this, we construct two paired features: (1) original $F^p{=}[\hat{f}^b;\hat{f}^a]$ and (2) reversed $F^r{=}[\hat{f}^a;\hat{f}^b]$, where the `before' and `after' features are swapped. Both are passed through the same type expert corresponding to the ground-truth change type, producing $F^m, F^{mr}\in\mathbb{R}^{h\times w\times 2d}$.
We normalize both along the feature dimension and compute cosine similarity between corresponding features. 
The expert consistency loss is defined based on change-type symmetry as:
\begin{equation}
    \mathcal{L}_{con} {=}
    \begin{cases}
        \displaystyle \frac{1}{B_{sym}} \sum_{i=1}^{B_{sym}} \left(1 - \cos(F^m_{i}, F^{mr}_{i})\right), & \text{if \textit{symmetry}}, \\
        \displaystyle \frac{1}{B_{asym}} \sum_{i=1}^{B_{asym}} \left(1 + \cos(F^m_{i}, F^{mr}_{i})\right), & \text{if \textit{asymmetry}},
    \end{cases}
\end{equation}
where \(B_{sym}\) and \(B_{asym}\) denote the number of samples for symmetric and asymmetric types, respectively.

Therefore, the loss enforces directional alignment between $F^m \text{ and } F^{mr}$ for symmetric types, while encouraging directional opposition for asymmetric types. By leveraging this type-specific orientation signal, each expert learns to encode the intrinsic characteristics that define its corresponding change type.

\subsection{Expert Disentangle Loss}
We propose an expert disentangle loss to explicitly separate the representation spaces of different change-type experts and the no-change expert. While the expert consistency loss guides each expert to capture type-intrinsic characteristics under input reversal, it does not prevent representations from different experts from becoming overly similar, particularly for visually correlated change types. The disentangle loss addresses this by promoting inter-type separability and isolating no-change features from change features. It consists of (1) an intra-group term that increases separation among change-type experts and (2) an inter-group term that separates change-type experts from the no-change expert.

For each expert \(t\), we compute a summary representation \(F^{m_t}_{avg}\) by applying GAP over its output feature \(F^{m_t}\).  
Let \(\mathcal{T}\) denote the set of change types, and \(T\) the no-change expert. The cosine similarity between any pair of expert representations is defined as \(d_{t,t'} = \cos(F^{m_t}_{avg}, F^{m_{t'}}_{avg})\).
The intra-group loss is defined as follows:
\begin{equation}
    \mathcal{L}_{intra} = 
    \frac{2}{|\mathcal{T}|(|\mathcal{T}| - 1)} \sum_{\substack{t, t' \in \mathcal{T} \\ t < t'}} \left[ \max\left(0, d_{t,t'} - \delta_{intra} \right) \right]^2.
\end{equation}

In addition, the inter-group loss is defined as:
\begin{equation}
    \mathcal{L}_{inter} = 
    \frac{1}{|\mathcal{T}|} \sum_{t \in \mathcal{T}} \left[ \max\left(0, d_{t,T} - \delta_{inter} \right) \right]^2,
\end{equation}
where \(\delta_{intra}\) and \(\delta_{inter}\) are margin hyperparameters for intra- and inter-group separation, respectively.
The total expert disentangle loss is computed as follows:
\begin{equation}
    \mathcal{L}_{dis} = \mathcal{L}_{intra} + \mathcal{L}_{inter}.
\end{equation}

Together, these terms explicitly prevent inter-type representation overlap, ensuring that each expert occupies a distinct region in the embedding space while maintaining clear separation from the no-change expert. A detailed figure is included in the supplementary material.

\subsection{Overall Objective}
The final training objective combines the captioning loss with all auxiliary terms:
\begin{equation}
    \mathcal{L}_{total} = \mathcal{L}_{cap} + \lambda_{1} \mathcal{L}_{router} + \lambda_{2} \mathcal{L}_{con} + \lambda_{3} \mathcal{L}_{dis},
\end{equation} 
where $\lambda_{1}, \lambda_{2}, \lambda_{3}$ are hyperparameters. $\mathcal{L}_{cap}$ denotes the captioning loss used to supervise the transformer decoder.

\section{Experiments}
\label{sec:Experiments}

\subsection{Datasets and Evaluation Metrics}
\noindent\textbf{Datasets.}
We conduct experiments on four datasets: CLEVR-DC \cite{clevrdc}, CLEVR-Change \cite{DUDA}, Spot-the-Diff \cite{spotthediff}, and Image Editing Request \cite{ier}. \textbf{CLEVR-DC} includes 48,000 image pairs, featuring more drastic scene differences due to strong distractors. We use the official split: 85\% for training, 5\% for validation, and 10\% for testing. \textbf{CLEVR-Change} contains 79,606 image pairs and 493,735 associated captions. We use the official split of 67,660/3,976/7,970 for train/val/test. \textbf{Spot-the-Diff} consists of 13,192 surveillance-based image pairs with illumination changes and uses a standard 8:1:1 split. \textbf{Image Editing Request (IER)} dataset comprises 3,939 image pairs and 5,695 editing instructions. We adopt the official split: 3,061 pairs for training, 383 for validation, and 495 for testing.

CLEVR-DC and CLEVR-Change follow a single change setting with six pre-defined change types (color, texture, add, drop, move, no-change). Spot-the-Diff supports both single- and multi-change settings and contains four types (add, drop, move, no-change). Image Editing Request also follows a single change setting with seven types (add, drop, replace, background, illumination, style, resize). The change-type labels for Spot-the-Diff and IER dataset are constructed using an GPT-4o \cite{hurst2024gpt} (see supplementary material for details). \\

\noindent\textbf{Evaluation Metrics.}
We evaluate the caption quality using five metrics: BLEU-4 ($\mathcal{B}$) \cite{bleu}, METEOR ($\mathcal{M}$) \cite{meteor}, ROUGE-L ($\mathcal{R}$) \cite{rouge}, CIDEr ($\mathcal{C}$) \cite{cider}, SPICE ($\mathcal{S}$) \cite{spice}.

\subsection{Implementation Details}
We apply MEDIC to representative recent change captioning models with publicly available code: SCORER \cite{scorer}, SMART \cite{smart}, and DIRL \cite{dirl}. We use $L{=}100$ memory slots. The number of experts is determined by the change-type taxonomy of each dataset. Specifically, we allocate one expert to each change type, including the no-change type. Thus, we use $T{=}6$ for CLEVR-DC and CLEVR-Change, $T{=}4$ for Spot-the-Diff, and $T=7$ for Image-Editing-Request. We set $\tau{=}0.1$, $\delta_{{intra}}{=}0.7$, and $\delta_{{inter}}{=}0.2$. The weights are $\lambda_1{=}0.1$, $\lambda_2{=}0.01$, and $\lambda_3{=}0.1$. All experiments use a single NVIDIA RTX 4090 GPU.

\subsection{Comparison under Single-Change Setting}
Table \ref{table:main_comparison} shows the performance comparison across three datasets, where MEDIC is applied to three baselines (SCORER \cite{scorer}, SMART \cite{smart}, DIRL \cite{dirl}). As SMART and DIRL lack CLEVR-DC and CLEVR-Change results respectively, and no method reports type-wise performance, we reproduced all missing results using publicly available official source code. Paired t-tests on all datasets confirm that the improvements of MEDIC over baselines are statistically significant ($p < 0.05$).\\

\noindent\textbf{Results on the CLEVR-DC Dataset.}
CLEVR-DC presents a particularly challenging setting, where severe viewpoint shifts act as strong distractors within the scene. As shown in Table \ref{table:main_comparison} (left), despite these challenges, applying MEDIC consistently yielded superior performance across all metrics. It demonstrates that explicitly separating change types during training is crucial for accurately capturing changes, especially in the presence of distractors. \\

\begin{table*}[t]
    \renewcommand{\tabcolsep}{1.0mm}
    \centering
    \caption{Comparison of change captioning performance on CLEVR-DC (left), CLEVR-Change (center), and Spot-the-Diff (right) datasets under single-change setting. Note that, our MEDIC is compared with three state-of-the-art baselines (SCORER~\cite{scorer}, SMART~\cite{smart}, and DIRL~\cite{dirl}) with publicly available source code.}
    \vspace{-0.25cm}
    \resizebox{\linewidth}{!}
    {
    \begin{tabular}{l | ccccc | ccccc | ccccc}
        \Xhline{3\arrayrulewidth}
        \multicolumn{1}{c|}{\rule{0pt}{9.3pt}\multirow{2}{*}[-0.1em]{\textbf{Method}}}
        & \multicolumn{5}{c|}{\cellcolor{gray!30}\textbf{CLEVR-DC Dataset}}
        & \multicolumn{5}{c|}{\cellcolor{gray!30}\textbf{CLEVR-Change Dataset}}
        & \multicolumn{5}{c}{\cellcolor{gray!30}\textbf{Spot-the-Diff Dataset}} \\
        \cline{2-6}\cline{7-11}\cline{12-16}
        \rule{0pt}{10.0pt} 
        & $\mathcal{B}$ & $\mathcal{M}$ & $\mathcal{R}$ & $\mathcal{C}$ & $\mathcal{S}$ 
        & $\mathcal{B}$ & $\mathcal{M}$ & $\mathcal{R}$ & $\mathcal{C}$ & $\mathcal{S}$ 
        & $\mathcal{B}$ & $\mathcal{M}$ & $\mathcal{R}$ & $\mathcal{C}$ & $\mathcal{S}$ \\\hline
            \rule{0pt}{10.0pt}DUDA (ICCV'19) \cite{DUDA} & 40.3 & 27.1 & - & 56.7 & 16.1 & 47.3 & 33.9 & - & 112.3 & 24.5 & - & - & - & - & - \\
            DUDA+ (CVPR'21) \cite{DUDA+} & - & - & - & - & - & 51.2 & 37.7 & 70.5 & 115.4 & 31.1 & 8.1 & 12.5 & - & 34.5 & - \\
            M-VAM (ECCV'20) \cite{M-VAM}& - & - & - & - & - & 50.3 & 37.0 & 69.7 & 114.9 & 30.5 & - & - & - & - & - \\
            VACC (ICCV'21) \cite{vacc} & 45.0 & 29.3 & - & 71.7 & 17.6 & - & - & - & - & - & - & - & - & - & - \\
            MCCFormers-D (ICCV'21) \cite{mccformer} & - & - & - & - & - & 52.4 & 38.3 & - & 121.6 & 26.8 & 10.0 & 12.4 & - & 43.1 & 18.3 \\
            MCCFormers-S (ICCV'21) \cite{mccformer} & - & - & - & - & - & 57.4 & 41.2 & - & 125.5 & 32.4 & - & 12.3 & - & 41.6 & 16.3 \\
            PCL w/o Pretrain (AAAI'22) \cite{PCL} & - & - & - & - & - & 32.7 & 27.7 & 57.2 & 89.8 & - & - & - & - & - & - \\
            NCT (TMM'23) \cite{NCT} & 47.5 & 32.5 & 65.1 & 76.9 & 15.6 & 55.1 & 40.2 & 73.8 & 124.1 & 32.9 & - & - & - & - & - \\
            VARD-Trans (TIP'23) \cite{vard} & 48.3 & 32.4 & - & 77.6 & 15.4 & 55.4 & 40.1 & 73.8 & 126.4 & 32.6 & - & 12.5 & 29.3 & 30.3 & 17.3 \\
            I3N-TD (TMM'23) \cite{i3n} & - & - & - & - & - & 55.8 & 40.6 & 73.9 & 125.6 & 32.8 & - & 13.0 & 31.5 & 42.7 & 18.6 \\
            RDD+ACR (AAAI'25) \cite{rdd} & - & - & - & - & - & 56.1 & 41.3 & 75.0 & 128.1 & 33.5 & 9.2 & 13.9 & 31.0 & 43.6 & - \\
            DECIDER (AAAI'25) \cite{decider} & - & - & - & - & - & 56.4 & 39.7 & 75.3 & 131.3 & - & 10.7 & 14.2 & 41.6 & 39.9 & - \\
            MCT-CCDiff (TIP'25) \cite{mct-ccdiff} & - & - & - & - & - & 57.5 & 40.6 & 75.6 & 131.7 & - & 10.8 & 14.5 & 35.5 & 41.7 & - \\\hline
            \rule{0pt}{10.0pt}SCORER (ICCV'23) \cite{scorer} & 49.4 & 33.4 & 66.1 & 83.7 & 16.2 & 56.3 & 41.2 & 74.5 & 126.8 & 33.3 & \textbf{10.2} & 12.2 & - & 38.9 & \textbf{18.4} \\
            \textbf{SCORER + MEDIC (Ours)} & \textbf{56.0} & \textbf{35.9} & \textbf{70.1} & \textbf{97.4} & \textbf{19.0} & \textbf{57.6} & \textbf{41.8} & \textbf{75.5} & \textbf{130.7} & \textbf{33.7} & \textbf{10.2} & \textbf{12.4} & \textbf{32.9} & \textbf{39.2} & \textbf{18.4} \\\hline
            
            \rule{0pt}{10.0pt}SMART (TPAMI'24) \cite{smart} & 48.3 & 30.7 & 65.2 & 81.2 & 16.0 & 56.1 & 40.8 & 74.2 & 127.0 & 33.4 & - & 13.5 & 31.6 & 39.4 & 19.0 \\
            \textbf{SMART + MEDIC (Ours)} & \textbf{57.1} & \textbf{35.3} & \textbf{70.9} & \textbf{98.9} & \textbf{19.8} & \textbf{56.4} & \textbf{42.5} & \textbf{75.9} & \textbf{128.1} & \textbf{34.6} & \textbf{9.1} & \textbf{14.2} & \textbf{32.6} & \textbf{43.1} & \textbf{22.1} \\\hline
            
            \rule{0pt}{10.0pt}DIRL (ECCV'24) \cite{dirl} & 51.4 & 32.3 & 66.3 & 84.1 & 16.8 & 56.2 & 41.0 & 73.8 & 126.0 & 33.1 & 10.3 & 13.8 & 32.8 & 40.9 & 19.9 \\
            \textbf{DIRL + MEDIC (Ours)} & \textbf{58.5} & \textbf{35.6} & \textbf{71.0} & \textbf{99.8} & \textbf{20.0} & \textbf{57.5} & \textbf{41.3} & \textbf{74.6} & \textbf{129.2} & \textbf{33.5} & \textbf{11.1} & \textbf{14.6} & \textbf{33.7} & \textbf{45.5} & \textbf{23.0} \\
            \Xhline{3\arrayrulewidth}
        \end{tabular}
    }
    \vspace{-0.3cm}
    \label{table:main_comparison}
\end{table*}

\noindent\textbf{Results on the CLEVR-Change Dataset.}
We also evaluate MEDIC on CLEVR-Change, which shares the same synthetic domain as CLEVR-DC but involves fewer distractors and more controlled scene variations. As shown in Table \ref{table:main_comparison} (center), MEDIC consistently improved performance across all evaluation metrics. Even in less cluttered environments, explicitly modeling change types contributes to better scene understanding and caption generation. \\

\noindent\textbf{Results on the Spot-the-Diff Dataset.}
In Table \ref{table:main_comparison} (right), we evaluate our method on Spot-the-Diff, a low-resolution real-world dataset with subtle illumination changes that make localization challenging. Despite these challenges, MEDIC consistently improves across all evaluation metrics, demonstrating strong robustness and generalization to complex and real-world scenarios.\\

\noindent\textbf{Type-wise Performance Comparison.}
We present a detailed comparison across change types in Table \ref{table:type_wise_comparison}, which contrasts the baseline DIRL with MEDIC-integrated model. The table includes six change types for CLEVR-DC and CLEVR-Change (`color', `texture', `add', `drop', `move', `no-change'), and four for Spot-the-Diff (`add', `drop', `move', `no-change'). MEDIC outperforms DIRL across all datasets and change types. Type-specific experts improve the ability of the model to capture diverse change patterns and achieve more accurate results.

\begin{table*}[t]
\centering

\begin{minipage}[t]{0.5\linewidth}
\centering
\renewcommand{\tabcolsep}{1.6mm}
\caption{CIDEr performance comparison by change type across datasets. Types: `Color' (Clr), `Texture' (Tex), `Add' (Add), `Drop' (Drp), `Move' (Mv), and `No-Change' (NC).}
\vspace{-0.21cm}
\resizebox{\linewidth}{!}{
\begin{tabular}{l | cccccc}
    \Xhline{3\arrayrulewidth}
    \multicolumn{7}{c}{\cellcolor{gray!30}\rule{0pt}{9.3pt}\textbf{CLEVR-DC Dataset}}\\\hline
    \multicolumn{1}{c |}{\rule{0pt}{10.0pt}\bf Method} & \textbf{Clr} & \textbf{Tex} & \textbf{Add} & \textbf{Drp} & \textbf{Mv} & \textbf{NC} \\\hline
    DIRL \cite{dirl} & 108.6 & 76.0 & 76.5 & 81.5 & 38.8 & 41.6\\
    \bf DIRL + MEDIC & \textbf{118.8} &  \textbf{88.2} & \textbf{86.1} & \textbf{88.3} & \textbf{54.4} & \textbf{82.3} \\\hline
    \multicolumn{7}{c}{\cellcolor{gray!30}\rule{0pt}{9.3pt}\textbf{CLEVR-Change Dataset}}\\\hline
    \multicolumn{1}{c |}{\bf Method} & \textbf{Clr} & \textbf{Tex} & \textbf{Add} & \textbf{Drp} & \textbf{Mv} & \textbf{NC} \\\hline
    DIRL \cite{dirl} & 149.8 & 137.5 & 126.9 & 137.5 & 87.6 & 113.8\\
    \bf DIRL + MEDIC & \textbf{151.7} & \textbf{142.6} & \textbf{138.6} & \textbf{141.2} & \textbf{96.3} & \textbf{116.2}\\\hline
    \multicolumn{7}{c}{\cellcolor{gray!30}\rule{0pt}{9.3pt}\textbf{Spot-the-Diff Dataset}}\\\hline
    \multicolumn{1}{c |}{\bf Method} & \textbf{Add} & \textbf{Drp} & \textbf{Mv} & \textbf{NC} & & \\\hline
    DIRL \cite{dirl} & 38.4 & 38.6 & 41.2 & 14.3 & & \\
    \bf DIRL + MEDIC & \textbf{47.2} & \textbf{47.6} & \textbf{48.4} & \textbf{17.4} & &\\
    \Xhline{3\arrayrulewidth}
\end{tabular}}
\vspace{-0.2cm}
\label{table:type_wise_comparison}
\end{minipage}
\hfill
\begin{minipage}[t]{0.49\linewidth}
\centering
\renewcommand{\tabcolsep}{1.8mm}
\caption{Comparison on the Spot-the-Diff under multi-change setting, following \cite{card}.}
\vspace{-0.35cm}
\resizebox{\linewidth}{!}{
\begin{tabular}{l | ccccc}
    \Xhline{3\arrayrulewidth}
    \multicolumn{1}{c |}{\bf Method} &
    $\mathcal{B}$ & $\mathcal{M}$ & $\mathcal{R}$ & $\mathcal{C}$ & $\mathcal{S}$ \\\hline
    SCORER \cite{scorer} & 5.1 & 9.3 & 23.0 & 20.9 & 11.9 \\
    \bf SCORER + MEDIC & \textbf{6.1} &  \textbf{9.3} & \textbf{25.7} & \textbf{25.9} & \textbf{16.4} \\\hline
    SMART \cite{smart} & 3.7 & 8.7 & 23.1 & 25.3 & 15.5 \\
    \bf SMART + MEDIC & \textbf{4.5} & \textbf{8.9} & \textbf{25.7} & \textbf{31.6} & \textbf{17.4} \\\hline
    DIRL \cite{dirl} & 4.6 & 9.5 & 23.8 & 21.5 & 14.8 \\
    \bf DIRL + MEDIC & \textbf{6.8} & \textbf{11.2} & \textbf{26.7} & \textbf{31.2} & \textbf{20.1}\\
    \Xhline{3\arrayrulewidth}
\end{tabular}}
\label{table:std_multi_change}

\vspace{9pt}
\centering
\captionof{table}{Expert architecture comparison of MEDIC on CLEVR-DC (DIRL baseline).}
\vspace{-0.35cm}
\renewcommand{\tabcolsep}{2.8mm}
\resizebox{\linewidth}{!}{
\begin{tabular}{c ccccc}
    \Xhline{3\arrayrulewidth}
    \rule{0pt}{10pt}\bf Architecture
     & $\mathcal{B}$ & $\mathcal{M}$ & $\mathcal{R}$ & $\mathcal{C}$ & $\mathcal{S}$\\\hline
    MLP (2 layer)
    & 56.1 & 35.1 & 70.1 & 96.1 & 19.3 \\
    Memory Network
    & \textbf{58.5} & \textbf{35.6} & \textbf{71.0} & \textbf{99.8} & \textbf{20.0} \\
    \Xhline{3\arrayrulewidth}
\end{tabular}}
\label{table:expert_architecture}
\vspace{-0.23cm}
\end{minipage}

\vspace{-0.2cm}
\end{table*}

\subsection{Comparison under Multi-Change Setting}

To verify the scalability of MEDIC to multi-change scenarios, we further conducted experiments on the Spot-the-Diff dataset~\cite{spotthediff}, the only publicly available dataset applicable to multi-change settings. While most previous works~\cite{decider, rdd} used the Spot-the-Diff dataset in a single-change setting, CARD~\cite{card} configured it for a multi-change setting and conducted experiments accordingly, which we followed. Using this protocol, we applied MEDIC to SCORER~\cite{scorer}, SMART~\cite{smart}, and DIRL~\cite{dirl}, reproducing the latter two for fair comparison. For this setting, we computed the consistency loss for each annotated type and took the mean across them, while using Binary Cross-Entropy for multi-label routing supervision. We generated multi-change type labels using a prompt detailed in the supplementary material. Table~\ref{table:std_multi_change} shows that MEDIC clearly outperforms all baselines, indicating that the soft gating mechanism effectively combines knowledge from multiple type experts and generalizes robustly beyond the single-change assumption.

\subsection{Ablation Studies}
To better understand the contribution of each component in our method, we conduct ablation studies on the CLEVR-DC, with MEDIC applied to DIRL \cite{dirl}. \\

\noindent\textbf{Effect of the Memory-based Expert.} 
To examine the impact of expert architecture within the same type-specific framework, we compare two expert designs: a standard MLP-based expert~\cite{MoE} and our memory expert.
Table \ref{table:expert_architecture} shows that ours outperforms the MLP-based expert across all metrics, verifying the advantages of input-adaptive retrieval in capturing type-specific cues.\\

\begin{figure*}[t]
\centering

\begin{minipage}[t]{0.48\linewidth}
\vspace{0pt}
\centering
\includegraphics[width=\linewidth]{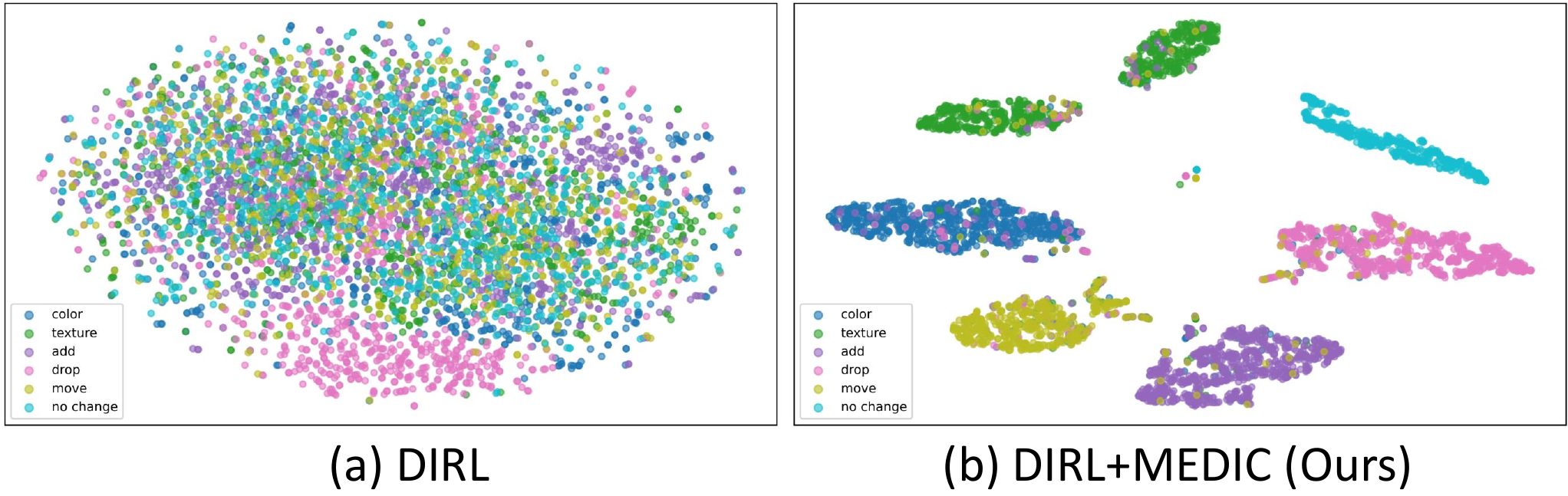}
\captionsetup{skip=3pt}
\vspace{-0.3cm}
\captionof{figure}{t-SNE visualization of learned representations by change type.}
\label{fig:t-sne}
\end{minipage}
\hfill
\begin{minipage}[t]{0.5\linewidth}
\centering
\vspace{-0.4cm}
\captionof{table}{Comparison between type-agnostic MoE and our type-specific expert.}
\label{table:standard_MoE}
\renewcommand{\tabcolsep}{2.5mm}
\resizebox{0.98\linewidth}{!}{
\begin{tabular}{ccc}
\toprule
\textbf{Method} & \textbf{\# Experts} & \textbf{CIDEr} \\
\midrule
\multirow{3}{*}{\begin{tabular}[c]{@{}l@{}}Standard MoE \\ (type-agnostic)\end{tabular}}
 & 6  & 94.2 \\
 & 12 & 90.7 \\
 & 24 & 93.6 \\\cdashline{1-3}
\rule{0pt}{10pt}DIRL+MEDIC (type-specific) & 6 & \textbf{99.8} \\
\bottomrule
\end{tabular}}
\end{minipage}

\end{figure*}

\noindent\textbf{t-SNE Visualization of Type-aware Experts.} 
To see whether each expert captures type-specific semantics, we visualize the expert outputs using t-SNE. As shown in Fig. \ref{fig:t-sne}, 
the baseline DIRL shows entangled clusters with unclear boundaries across change types. In contrast, MEDIC generates well-separated clusters, which shows that each expert learns more discriminative and type-specific representations and highlights the effectiveness of type-aware modeling. \\

\begin{table*}[t]
    \centering

    \begin{minipage}[t]{0.46\linewidth}
        \vspace{0pt}
        \centering
        \renewcommand{\tabcolsep}{1.8mm}
        \caption{Effect of proposed losses on the CLEVR-DC dataset.}
        \vspace{-0.3cm}
        \resizebox{\linewidth}{!}{
        \begin{tabular}{c ccccc}
            \Xhline{3\arrayrulewidth}
            \rule{0pt}{10pt}\textbf{Settings} & $\mathcal{B}$ & $\mathcal{M}$ & $\mathcal{R}$ & $\mathcal{C}$ & $\mathcal{S}$ \\\hline
            \rule{0pt}{9.3pt}DIRL + MEDIC & \textbf{58.5} & \textbf{35.6} & \textbf{71.0} & \textbf{99.8} & \textbf{20.0} \\
            \cdashline{1-6}
            \rule{0pt}{9pt}w/o $\mathcal{L}_{dis}$ & 57.7 & 35.2 & 70.6 & 98.2 & 19.3 \\
            w/o $\mathcal{L}_{con}$ & 57.0 & 35.0 & 70.5 & 99.1 & 19.7 \\
            w/o $\mathcal{L}_{dis}$, $\mathcal{L}_{con}$ & 54.8 & 34.9 & 69.9 & 96.7 & 18.4 \\
            w/o $\mathcal{L}_{router}$, $\mathcal{L}_{dis}$, $\mathcal{L}_{con}$ & 53.2 & 33.0 & 69.9 & 87.7 & 18.5 \\
            \Xhline{3\arrayrulewidth}
        \end{tabular}}
        \label{table:loss_ablation}
    \end{minipage}
    \hfill
    \begin{minipage}[t]{0.53\linewidth}
    \vspace{0pt}
        \centering
        \renewcommand{\tabcolsep}{0.8mm}
        \caption{Ablation on the number of memory slots per expert on CLEVR-DC dataset.}
        \vspace{-0.3cm}
        \resizebox{\linewidth}{!}{
        \begin{tabular}{c ccccc | c c}
            \Xhline{3\arrayrulewidth}
            \rule{0pt}{9.5pt}\bf \# Slot
             & $\mathcal{B}$ & $\mathcal{M}$ & $\mathcal{R}$ & $\mathcal{C}$ & $\mathcal{S}$ & \# Params(M) & Time(ms/sample)\\\hline
            \rule{0pt}{9.3pt}-
            & 51.4 & 32.3 & 66.3 & 84.1 & 16.8 & 13.90 & 26.90 \\\cdashline{1-8}
            \rule{0pt}{9.2pt}50
            & 56.6 & 35.2 & 70.5 & 97.8 & 19.0 & 24.50 & 30.11  \\
            \textbf{100}
            & \textbf{58.5} & \textbf{35.6} & \textbf{71.0} & \textbf{99.8} & \textbf{20.0} & \textbf{25.12} & \textbf{30.16} \\
            200
            & 57.8 & 35.1 & 70.3 & 98.8 & 18.2 & 26.35 & 30.21 \\
            400
            & 56.9 & 35.0 & 70.9 & 97.7 & 18.5 & 28.80 & 30.22 \\
            \Xhline{3\arrayrulewidth}
        \end{tabular}}
        \label{table:slot_ablation}
    \end{minipage}

\vspace{-0.3cm}
\end{table*}

\begin{figure}[t]
    \centering
    \includegraphics[width=0.95\linewidth]{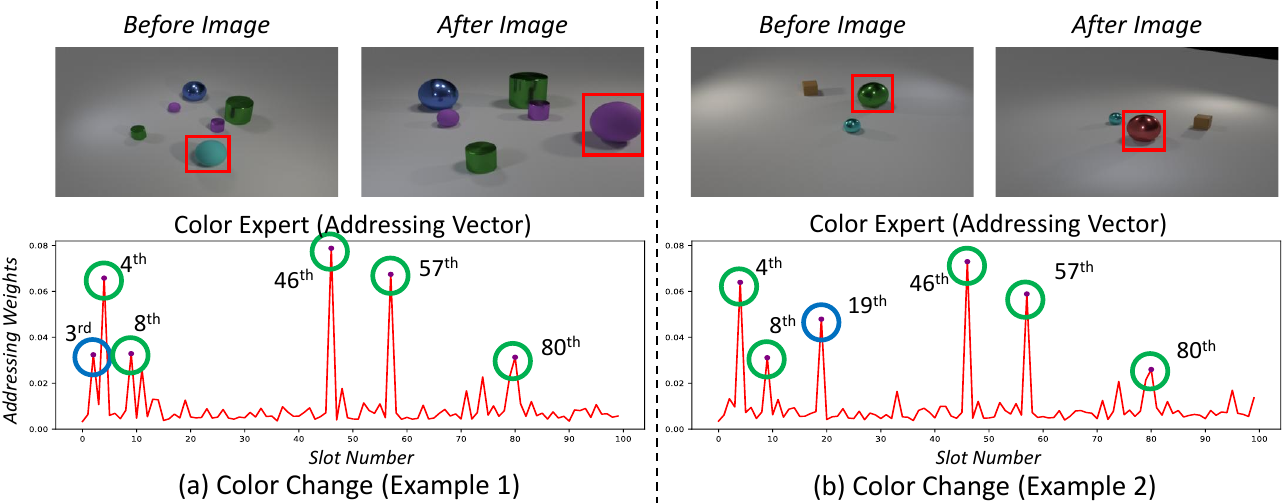}
    \vspace{-0.1cm}
    \caption{Visualization of two image pairs with the same change type (color), and their slot activation profiles. Green circles indicate commonly activated slots across inputs, capturing type-specific patterns, while blue circles highlight input-specific activations.}
    \label{fig:slot_activation}
\end{figure}

\noindent\textbf{Effectiveness of the Type-Specific Expert Design.}
To validate the effectiveness of our type-specific expert design, we compare MEDIC with a standard type-agnostic MoE. As shown in Table~\ref{table:standard_MoE}, our design consistently achieves higher performance across different expert configurations. This result indicates that a type-specific expert design is more effective than a generic MoE design. We further compare MEDIC with a simple type-aware baseline trained with an auxiliary change-type classification loss in the supplementary material, showing that the gains achieved by MEDIC are not merely due to access to type labels but come from explicit type-specialized expert reasoning.\\

\noindent\textbf{Effect of the Proposed Losses.} 
We analyze the impact of the expert consistency loss $\mathcal{L}_{con}$, expert disentangle loss $\mathcal{L}_{dis}$, and routing loss $\mathcal{L}_{router}$ (Table~\ref{table:loss_ablation}). Removing $\mathcal{L}_{dis}$ or $\mathcal{L}_{con}$ reduces performance, while each individually still improves results, indicating their complementary role of expert specialization. Removing $\mathcal{L}_{router}$ forces implicit expert selection and prevents explicit routing training, and leads to further performance drops. Combining all three losses achieves the best performance and demonstrates their synergy in learning type-specialized experts.\\

\noindent\textbf{Effect of the Memory Slot Size and Overhead.}
We analyze how the number of memory slots per expert affects the results and computational cost. As shown in Table \ref{table:slot_ablation}, increasing the number of slots improves performance up to 100, achieving the best results with only a marginal cost increase. Although performance slightly drops beyond this point, it still surpasses the non-memory baseline.

\subsection{Discussion}

\noindent\textbf{Dynamic Slot Activation.}
To verify that our memory-based experts adapt to input content, we visualize the addressing vectors of two color-type samples in Fig.~\ref{fig:slot_activation}. Both consistently attend to specific slots (green circles), indicating anchors for type-specific information, while distinct slots (blue circles) reflect input-specific responses. This shows that our memory architecture captures both shared patterns and adaptive behavior within each change type.\\

\noindent\textbf{Effect of the Hyper-parameters.} As shown in Table~\ref{table:lambda_ablation}, MEDIC remains stable across a wide range of hyper-parameters $\lambda_1$, $\lambda_2$, and $\lambda_3$ values, with all configurations achieving strong results and only minor variation. \\

\noindent\textbf{Generalization to Object- and Global-Level Changes.} 
To see that MEDIC can extend beyond object-level changes, we evaluate it on the Image Editing Request (IER) dataset~\cite{ier} that contains more diverse change scenarios. While most existing datasets primarily focus on object-level changes, IER includes three object-level types (`add', `drop', `replace') and four global-level types (`background', `illumination', `style', `resize'). As shown in Table~\ref{table:IER}, when using DIRL as the baseline, MEDIC achieves consistent improvements across all change types. It demonstrates robustness across both object-level and global changes. \\ 

\begin{table*}[t]
\centering

\begin{minipage}[t]{0.52\linewidth}
\vspace{0pt}
\centering
\renewcommand{\tabcolsep}{2.5mm}
\caption{Evaluation results under different settings of the hyper-parameters on the CLEVR-DC dataset.}
\vspace{-0.3cm}
\resizebox{\linewidth}{!}{
\begin{tabular}{c|c|ccccc}
    \Xhline{3\arrayrulewidth}
    \textbf{Target} & $\boldsymbol{\lambda}$ & $\mathcal{B}$ & $\mathcal{M}$ & $\mathcal{R}$ & $\mathcal{C}$ & $\mathcal{S}$ \\
    \hline
    \multirow{3}{*}{$\lambda_1$}
        & \textbf{0.1} & \textbf{58.5} & \textbf{35.6} & \textbf{71.0} & \textbf{99.8} & \textbf{20.0} \\
        & 0.01 & 58.1 & 35.5 & 71.1 & 99.2 & 19.3 \\
        & 0.001 & 56.5 & 35.6 & 70.7 & 98.6 & 18.6 \\
    \hline
    \multirow{3}{*}{$\lambda_2$}
        & 0.1 & 58.2 & 35.5 & 71.1 & 99.3 & 19.5 \\
        & \textbf{0.01} & \textbf{58.5} & \textbf{35.6} & \textbf{71.0} & \textbf{99.8} & \textbf{20.0} \\
        & 0.001 & 57.4 & 35.3 & 70.8 & 99.6 & 19.5 \\
    \hline
    \multirow{3}{*}{$\lambda_3$}
        & \textbf{0.1} & \textbf{58.5} & \textbf{35.6} & \textbf{71.0} & \textbf{99.8} & \textbf{20.0} \\
        & 0.01 & 57.7 & 35.6 & 70.9 & 98.9 & 19.9 \\
        & 0.001 & 57.4 & 35.3 & 70.6 & 98.2 & 20.0 \\
    \Xhline{3\arrayrulewidth}
\end{tabular}}
\vspace{-0.4cm}
\label{table:lambda_ablation}
\end{minipage}
\hfill
\begin{minipage}[t]{0.46\linewidth}
\vspace{0pt}
\centering

\renewcommand{\tabcolsep}{2.3mm}
\captionof{table}{Results on IER dataset.}
\vspace{-0.2cm}
\resizebox{\linewidth}{!}{
\begin{tabular}{lcccc}
\toprule
\textbf{Method} & $\mathcal{B}$ & $\mathcal{M}$ & $\mathcal{R}$ & $\mathcal{C}$ \\
\midrule
DIRL \cite{dirl} & 10.9 & 15.0 & 41.0 & 34.1 \\
DIRL+MEDIC & \textbf{11.2} & \textbf{15.6} & \textbf{42.5} & \textbf{37.2} \\
\bottomrule
\end{tabular}}
\label{table:IER}

\vspace{10pt}

\renewcommand{\tabcolsep}{2.2mm}
\captionof{table}{Results on LVLM-based works.}
\vspace{-0.4cm}
\resizebox{\linewidth}{!}{
\begin{tabular}{lccc}
\toprule
\textbf{Method} & $\mathcal{B}$ & $\mathcal{M}$ & $\mathcal{C}$ \\
\midrule
BLIP2IDC \cite{blip2idc} & 49.3 & 33.0 & 88.5 \\
BLIP2IDC+MEDIC & \textbf{57.3} & \textbf{33.8} & \textbf{106.4} \\
\bottomrule
\end{tabular}}
\vspace{-0.5cm}
\label{table:BLIP2IDC}
\end{minipage}
\vspace{-0.02cm}
\end{table*}

\noindent\textbf{Compatibility with LVLM-Based Methods.}
To further verify compatibility with recent LVLM-based approaches, we apply MEDIC to BLIP2IDC\cite{blip2idc} and evaluate it on CLEVR-DC. As shown in Table~\ref{table:BLIP2IDC}, incorporating MEDIC consistently improves performance across all metrics, demonstrating that our method can be effectively integrated with LVLM-based models. Additional comparisons with zero-shot and fine-tuned LVLM-based change captioning methods are provided in the supplementary material.\\

\noindent\textbf{Label Reliability \& Routing Robustness.}
For datasets without type annotations, we derive change-type labels from ground-truth captions using GPT-4o. To validate this supervision, applying the same labeling protocol to CLEVR-Change and CLEVR-DC, where ground-truth type labels are available, yields 91.2\% and 93.8\% accuracy, respectively. Moreover, even when 10/20/40\% of CLEVR-DC training type labels are randomly corrupted, MEDIC achieves 95.4/94.3/94.0 CIDEr, still outperforming DIRL (84.1), showing robustness to imperfect type supervision. In addition, stage-1/2 routing accuracies show 92\%/81\%, respectively, and qualitative routing-error analysis shows that our soft expert aggregation can compensate for some top-1 routing mistakes. Detailed prompts, label validation, noisy-label analysis, and routing-error examples are provided in the supplementary material.\\

\noindent\textbf{Generalization to In-Domain Held-Out Types }
As MEDIC is built on a modular design with dedicated experts for each change type, it can be extended when additional supervised change types become available. To examine this property, we conduct a leave-one-type-out evaluation on CLEVR-DC in an in-domain held-out setting: one existing change type, such as `color', is excluded during training and evaluated afterward.
When `color' is held out, CIDEr on that type drops from 118.8 to 18.4.
When this held-out type is later introduced, MEDIC can
add a new expert and update the expanded router while freezing the existing experts. With only 0.2M additional parameters (0.82\% of the 25.12M total), CIDEr on the held-out type recovers from 18.4 to 113.4 while largely preserving performance on previously learned types. Details are provided in the supplementary material.\\

\noindent\textbf{Limitations.}
While MEDIC can flexibly extend to unseen datasets by using LLMs to extract new change-type labels, it still relies on the availability of such type annotations. Future work will explore self-supervised approaches that can infer change types without explicit labels.

\begin{figure*}[t]
    \centering
    \includegraphics[width=\linewidth]{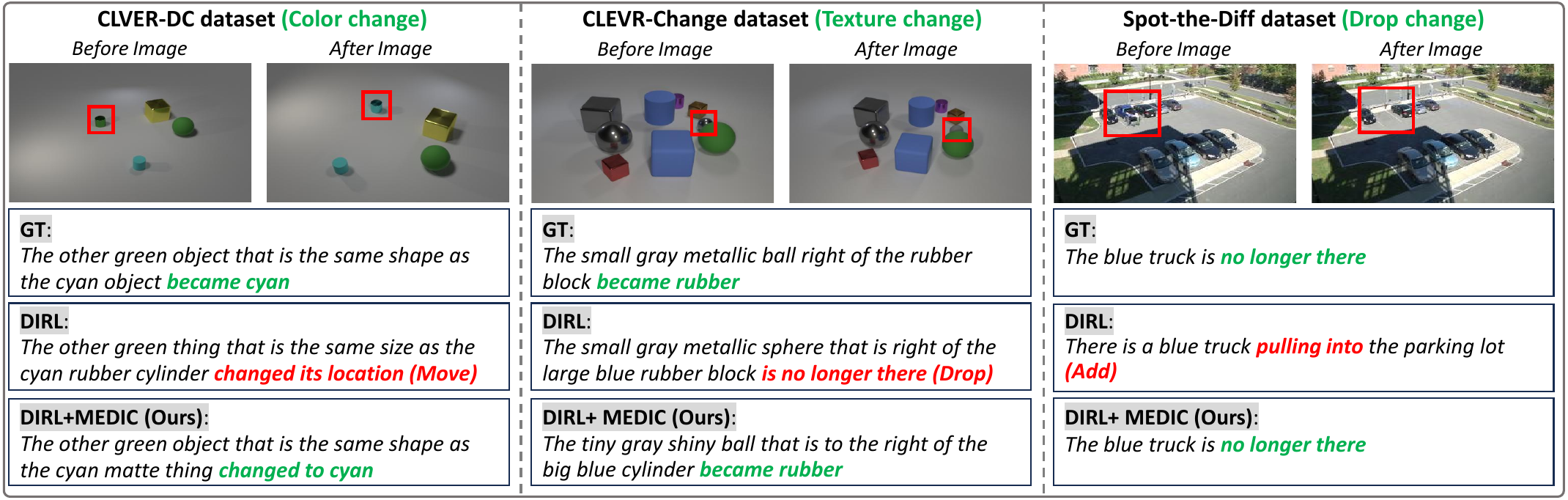}
    \vspace{-0.6cm}
    \caption{Qualitative comparisons across datasets. 
    Each example shows ground-truth (GT), baseline (DIRL), and the proposed method (DIRL+MEDIC).}
    \vspace{-0.45cm}
    \label{fig:qualitative_results}
\end{figure*}

\subsection{Qualitative Results}

Figure~\ref{fig:qualitative_results} presents qualitative examples on CLEVR-Change, CLEVR-DC, Spot-the-Diff, and Image-Editing-Request. MEDIC generally produces more accurate and type-consistent captions than DIRL by explicitly modeling change types. We additionally provide representative failure cases and their analysis in the supplementary material to highlight the remaining limitations of MEDIC under subtle local changes, crowded multi-change scenes, and ambiguous global edits.

\section{Conclusion}
We introduce MEDIC, a novel framework that explicitly models change types using memory-based experts specialized for the change type. These experts dynamically retrieve type-relevant visual patterns to enable input-adaptive and type-aware reasoning for accurate change descriptions. Extensive experiments show that explicit change-type modeling leads to consistent performance across diverse and challenging scenarios.

\section*{Acknowledgements} This work was partly supported by IITP-ITRC grant funded by the Korea government (MSIT)(IITP-2026-RS-2023-00258649, 30\%) and partly supported by IITP grant funded by the Korea government (MSIT)(IITP-2023-RS-2023-00266615: Convergence Security Core Talent Training Business Support Program (20\%), IITP-2022-II220078: Explainable Logical Reasoning for Medical Knowledge Generation (25\%), No. RS-2024-00509257: Global AI Frontier Lab (25\%)).

%
%
\bibliographystyle{splncs04}
\bibliography{main}

\clearpage
 \title{Different Changes Require Different Reasoning: Change-Type-Specialized Experts for Robust Change Captioning \\
-- \textit{Supplementary Material} --}

\titlerunning{MEDIC}

\author{
Jiyoung Park\equalcontrib\orcidlink{0009-0000-4212-7745}
\and
InJae Oh\equalcontrib\orcidlink{0009-0001-1321-3019}
\and
Jung Uk Kim\corrauthor\orcidlink{0000-0003-4533-4875}
}

\authorrunning{J. Park et al.}

\institute{
Kyung Hee University, Yong-in, South Korea \\
\email{\{jy0117, seanoh, ju.kim\}@khu.ac.kr}
}

\maketitle

\blfootnote{\textsuperscript{*} Equal contribution. \quad
\textsuperscript{$\dagger$} Corresponding author.}

\section*{Overview}
In this supplementary material, we present additional analyses and results that complement the main paper. Specifically, we provide:

\begin{tcolorbox}[colback=gray!5!white, colframe=gray!75!black, title=Table of Contents]
\begin{enumerate}[label=\Alph*., leftmargin=*, itemsep=2pt]
    \item {Explanation of Expert Disentangle Loss} \dotfill \pageref{sec:disentangle loss}
    \item {Prompting and Validation of LLM-Generated Type Labels} \dotfill \pageref{sec:prompt}
    \item {{Robustness to Noisy Type Labels}} \dotfill \pageref{sec:noisy_label}
    \item {Analysis of the Relationship Between Router Predictions and Generated Captions} \dotfill \pageref{sec:router}
    \item {Type-Aware Baseline Analysis} \dotfill \pageref{sec:type_aware_baseline}
    \item {Leave-One-Type-Out Evaluation for In-Domain Held-Out Change Types} \dotfill \pageref{sec:held_out_type}
    \item {Taxonomy-Mismatch Stress Test} \dotfill \pageref{sec:taxonomy-mismatch}
    \item {Impact of Routing Accuracy on Caption Quality} \dotfill \pageref{sec:routing acc and caption quality}
    \item {Ablation of Routing Strategy} \dotfill \pageref{sec:routing_strategy}
    \item {Additional Evaluation Beyond Caption Similarity Metrics} \dotfill \pageref{sec:detailed_evaluation}
    \item {Comparison with LVLM-Based Methods} \dotfill \pageref{sec:mllm_comparison}
    \item {Qualitative Results} \dotfill \pageref{sec:qualitative results}
    \item {Failure Cases} \dotfill \pageref{sec:failure cases}
    \item {Dynamic Slot Activation Analysis} \dotfill \pageref{sec:dynamic slot activation}
\end{enumerate}
\end{tcolorbox}

This document offers in-depth qualitative and quantitative analyses to support the effectiveness of our method. The included figures, ablations, and visualizations provide further understanding of how change-type awareness and memory-based experts contribute to improved change captioning.

\begin{figure*}[t!]
    \begin{minipage}[b]{1\linewidth}
	\centering
        \centerline{\includegraphics[width=\linewidth]{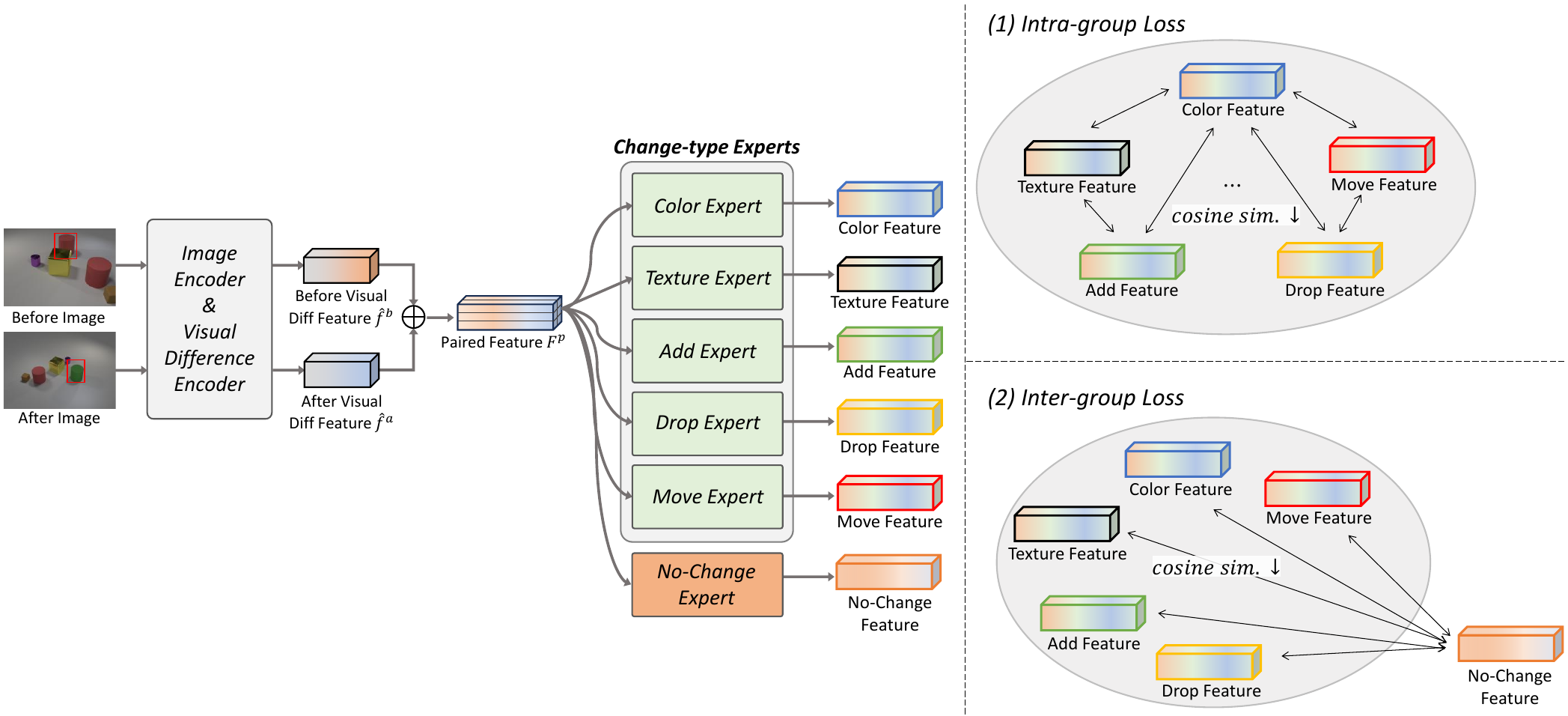}}
        \end{minipage}
    \centering
    \vspace{-0.1cm}
	\caption{Conceptual illustration of the expert disentangle loss. 
    Within the change group (top-right), different type-specific experts are encouraged to stay apart from one another. 
    Across groups (bottom-right), each change expert is pushed away from the no-change expert, enforcing pairwise inter-group separation between all change/no-change expert pairs to ensure semantic separation.}
    \label{fig:disentangle_visualization}
    \vspace{-0.3cm}
\end{figure*}

\section{Explanation of Expert Disentangle Loss}
\label{sec:disentangle loss}
As introduced in the ``\textbf{Expert Disentangle Loss}'' section of the main paper (Section 3.5), this section provides a detailed explanation of how the loss operates, along with a conceptual visualization for better understanding. The expert disentangle loss aims to organize the representation space by enforcing separation among change-type experts and distinguishing them from the no-change expert. It is composed of two components: (1) an intra-group loss that enforces disentanglement among change-type experts, and (2) an inter-group loss that separates all change-type experts from the no-change expert. \\

\noindent\textbf{Feature Construction.}
\begin{itemize}
    \item As illustrated in Figure~\ref{fig:disentangle_visualization}, for each input image pair, visual difference features from the before and after images, denoted as \( \hat{f}^b \) and \( \hat{f}^a \), respectively. These are concatenated to form the paired feature \( F^p = [\hat{f}^b; \hat{f}^a] \), which serves as the input to all experts.
    \item The paired feature is routed to the corresponding expert based on the ground-truth change type. Each expert produces a type-specific output feature \( F^{m_t} \), which captures visual evidence relevant to its assigned change type (\textit{e.g.,} color, texture).
    \item Global average pooling (GAP) is applied to each output of the expert to obtain a summary representation \( F^{m_t}_{avg} \in \mathbb{R}^{2d} \), which is used to compute similarity among expert outputs. \\
\end{itemize}

\noindent\textbf{Loss Mechanism.}
\begin{itemize}
    \item \textbf{Intra-group Loss} (Figure~\ref{fig:disentangle_visualization} (top-right)): To ensure distinct representations among change-type experts, the intra-group loss penalizes high similarity between different change-type experts. For every pair of change-type experts \( (t, t') \), the cosine similarity \( d_{t,t'} = \cos(F^{m_t}_{avg}, F^{m_{t'}}_{avg}) \) is computed. The loss term is defined as:
    \[
    \mathcal{L}_{\text{intra}} = \frac{2}{|\mathcal{T}|(|\mathcal{T}| - 1)} \sum_{\substack{t, t' \in \mathcal{T} \\ t < t'}} \left[ \max\left(0, d_{t,t'} - \delta_{\text{intra}} \right) \right]^2
    \]
    where \( \mathcal{T} \) is the set of change types, and \( \delta_{\text{intra}} \) is a margin that softly enforces a minimum dissimilarity among experts.\\

    \item \textbf{Inter-group Loss} (Figure~\ref{fig:disentangle_visualization} (bottom-right)): The inter-group loss separates the change-type experts from the no-change expert. For each change-type expert \( t \in \mathcal{T} \), the similarity with the no-change expert \( T \) is computed as \( d_{t,T} = \cos(F^{m_t}_{avg}, F^{m_T}_{avg}) \), and the loss is defined as:
    \[
    \mathcal{L}_{\text{inter}} = \frac{1}{|\mathcal{T}|} \sum_{t \in \mathcal{T}} \left[ \max\left(0, d_{t,T} - \delta_{\text{inter}} \right) \right]^2
    \]

    where \( \delta_{\text{inter}} \) is a margin that enforces separation between changed and unchanged representations. \\
 
    \item \textbf{Total Disentangle Loss:}  
    The overall loss is the sum of the two terms:
    \[
    \mathcal{L}_{\text{dis}} = \mathcal{L}_{\text{intra}} + \mathcal{L}_{\text{inter}} \\
    \]
\end{itemize}

\noindent\textbf{Interpretation.}
Figure~\ref{fig:disentangle_visualization} shows a conceptual illustration of the expert disentangle loss. Type-specific experts are encouraged to diverge within the change group while being separated from the no-change expert, resulting in a well-structured and disentangled representation space.






\section{Prompting and Validation of LLM-Generated Type Labels}
\label{sec:prompt}
In this section, we present our approach for generating change-type labels for the Spot-the-Diff \cite{spotthediff} and Image-Editing-Request \cite{ier} datasets using GPT-4o \cite{hurst2024gpt}, including the detailed prompting strategy and an analysis of label quality, as mentioned in the Dataset and Discussion section of the main paper. Since both datasets lack ground-truth type labels, we address this limitation by generating them through large language model prompting.

We utilize the ground-truth captions provided in the dataset, which describe the semantic differences between the two images. Since these captions inherently contain information about what has changed, they serve as a strong signal for inferring change types. We prompt GPT-4o with the caption and ask it to predict the corresponding change type. The full prompts for the single-change and multi-change scenarios are shown in Table~\ref{prompt_single_std}, Table~\ref{prompt_single_ier} and Table~\ref{prompt_multi}, respectively. We adopt in-context learning to guide the predictions of model.

In single-change scenarios, to evaluate the quality of the LLM-generated type labels, we apply the same prompting strategy to two benchmark datasets—CLEVR-Change \cite{DUDA} and CLEVR-DC \cite{clevrdc}—which contain ground-truth type labels. By comparing the LLM-predicted labels to the ground-truth type labels, we assess the accuracy of LLM. As shown in Table~\ref{table:llm_validation}, the accuracy is high on both datasets, demonstrating that our method produces high-quality type labels even in scenarios where ground-truth annotations are unavailable, such as real-world datasets. In the multi-change setting, we do not conduct a similar evaluation because Spot-the-Diff \cite{spotthediff} is the only publicly available multi-change dataset.

 {While the above validation confirms that the LLM-generated labels are generally reliable, some label noise may still exist in datasets without ground-truth type annotations. We therefore further analyze the robustness of MEDIC to corrupted type supervision in the following section.}

\begin{table}[t]
\centering

\begin{minipage}[t]{0.38\linewidth}
\centering
\renewcommand{\tabcolsep}{2.3mm}
\caption{Validation of type labels extracted using the Spot-the-Diff approach on CLEVR-Change and CLEVR-DC, both with ground-truth types.}
\vspace{1mm}
\resizebox{0.95\linewidth}{!}{
    \begin{tabular}{lc}
    \toprule
    \rule{0pt}{8pt}\textbf{Dataset} & \textbf{Accuracy(\%)} \\
    \midrule
    CLEVR-Change & 91.2 \\
    CLEVR-DC & 93.8 \\
    \bottomrule
    \end{tabular}}
\label{table:llm_validation}
\end{minipage}
\hfill
\begin{minipage}[t]{0.58\linewidth}
\centering
\renewcommand{\tabcolsep}{2.5mm}
\caption{Results on noisy type labels. We randomly corrupt a portion of training type labels on CLEVR-DC while keeping image pairs and caption supervision unchanged.}
\vspace{1mm}
\resizebox{0.95\linewidth}{!}{
\begin{tabular}{lcc}
\toprule
\textbf{Method} &
\textbf{Type Label Noise} &
\textbf{CIDEr} \\
\midrule

\multirow{4}{*}{{DIRL+MEDIC}}
& {0\%}  & \textbf{ {99.8}} \\
& {10\%} &  {95.4 (-4.4)} \\
& {20\%} &  {94.3 (-5.5)} \\
&  {40\%} &  {94.0 (-5.8)} \\

\midrule

 {DIRL} &
 {N/A} &
 {84.1} \\

\bottomrule
\end{tabular}}
\label{table:noisy_label}
\end{minipage}

\end{table}

\section{ {Robustness to Noisy Type Labels}}
\label{sec:noisy_label}

 {LLM-generated type labels on Spot-the-Diff and Image-Editing-Request may contain potential noise, since these datasets do not provide ground-truth change-type annotations. To examine whether MEDIC is overly dependent on perfectly clean type supervision, we conduct a noisy-label stress test on CLEVR-DC, where ground-truth type labels are available.}

 {Specifically, we randomly corrupt 10\%, 20\%, and 40\% of the training type labels by replacing each selected label with an incorrect label from the change-type taxonomy. The image pairs and caption supervision are kept unchanged, so only the type-level supervision used for routing and expert specialization is corrupted.}

 {As shown in Table~\ref{table:noisy_label}, MEDIC achieves 95.4, 94.3, and 94.0 CIDEr under 10\%, 20\%, and 40\% type-label noise, respectively. Although performance decreases as the noise ratio increases, MEDIC consistently remains above the DIRL baseline, which achieves 84.1 CIDEr. Even under 40\% noisy type labels, MEDIC still preserves a substantial improvement over DIRL.}

 {These results indicate that MEDIC does not rely solely on discrete type labels. Instead, its expert routing and caption generation are jointly guided by image-pair features, caption supervision, and soft expert aggregation. 

\begin{figure}[t]
    \begin{minipage}[b]{0.999\linewidth}
	\centering
        \centerline{\includegraphics[width=\linewidth]{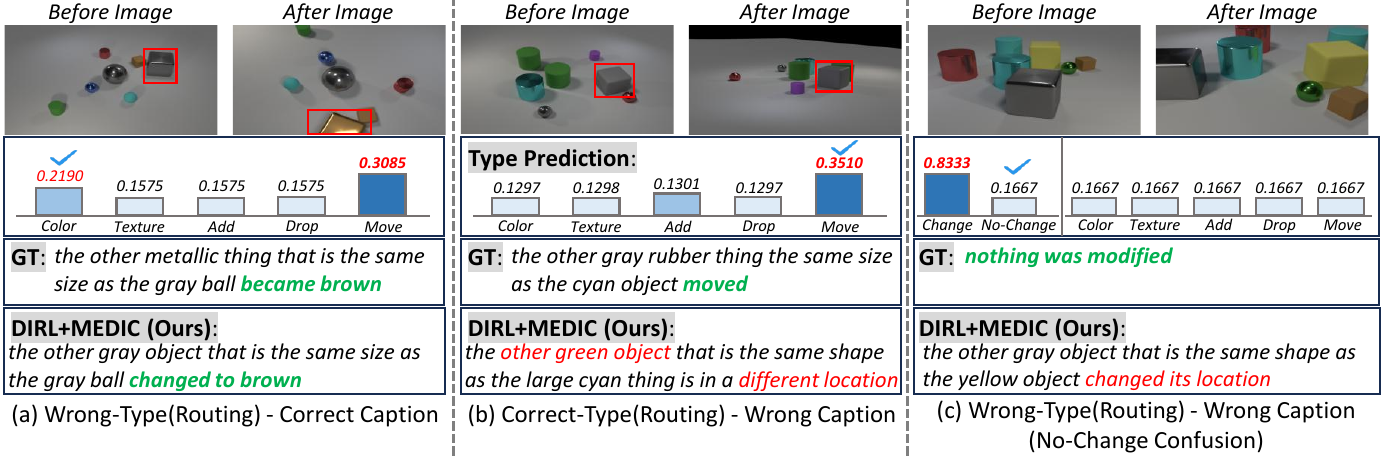}}
        \end{minipage}
    \centering
	\caption{Visualization of MEDIC under imperfect routing.}
    \label{fig:imperfect routing}
\end{figure}

\section{Analysis of the Relationship Between Router Predictions and Generated Captions}
\label{sec:router}
To better understand behavior of MEDIC under imperfect routing, we conduct a qualitative analysis on CLEVR-Change, CLEVR-DC, and Spot-the-Diff (Figure~\ref{fig:imperfect routing} (a), Figure~\ref{fig:router_analysis}), focusing on challenging cases where the router’s top-1 predicted change type differs from the ground-truth (GT). Despite this mismatch, the generated captions often correctly describe both the changed object and the modification itself, indicating that a non-GT top-1 prediction does not necessarily lead to captioning failure.

This robustness stems from the soft expert aggregation mechanism: even when the GT type is not ranked first, it typically receives the second-largest probability with a value close to the top-1 score. As MEDIC aggregates all type-specific experts through weighted summation, the GT expert still contributes substantially, allowing the model to integrate information from multiple plausible change types. The bar plots in Figure~\ref{fig:router_analysis} illustrate this behavior, where the GT probability remains non-negligible even when it is not the highest.

Such routing ambiguities arise for different reasons across datasets. In CLEVR-Change and CLEVR-DC, subtle modifications (\textit{e.g.,} small objects, partial occlusions, or minor displacements) can blur the distinction between visually related types, causing probability mass to spread across competing categories. In Spot-the-Diff, a single image pair may contain multiple change types, while the router predicts only one label, making the top-1 prediction reflect a dominant pattern rather than all underlying changes. Nevertheless, soft aggregation enables MEDIC to maintain reliable caption generation in many of these cases.

In addition to the mitigation cases discussed above, Figure~\ref{fig:imperfect routing} (b), (c) also present representative failure cases. In Figure~\ref{fig:imperfect routing} (b), the router predicts the correct type but the caption is inaccurate, showing that correct routing alone does not guarantee accurate description. In Figure~\ref{fig:imperfect routing} (c), both routing and captioning fail, including no-change confusion. These failures typically occur under extreme viewpoint variations or substantial scene reconfiguration, where unreliable object alignment between the before and after images leads to ambiguous correspondences that adversely affect the final caption.

\begin{table}[t]
\renewcommand{\tabcolsep}{2.3mm}
\centering
\caption{ {Comparison with a simple type-aware baseline on CLEVR-DC using DIRL as the backbone. ``Aux. Type Loss'' denotes adding an auxiliary change-type classification loss to DIRL without using type-specific experts.}}
\resizebox{0.9\linewidth}{!}{
\begin{tabular}{lccc}
\toprule
\textbf{ {Method}} &
\textbf{ {Type Labels}} &
\textbf{ {Type-Specific Experts}} &
\textbf{ {CIDEr}} \\
\midrule
 {DIRL} &
 {\xmark} &
 {\xmark} &
 {84.1} \\
 {DIRL + Aux. Type Loss} &
 {\cmark} &
 {\xmark} &
 {87.2} \\
 {DIRL + MEDIC} &
 {\cmark} &
 {\cmark} &
\textbf{ {99.8}} \\
\bottomrule
\end{tabular}}
\label{table:type_aware_baseline}
\end{table}

\section{ {Type-Aware Baseline Analysis}}
\label{sec:type_aware_baseline}

 {To verify that the improvement of MEDIC does not simply come from access to change-type labels, we compare MEDIC with a simpler type-aware baseline. Specifically, we train DIRL on CLEVR-DC with an auxiliary change-type classification loss while keeping the captioning architecture type-agnostic.}

 {As shown in Table~\ref{table:type_aware_baseline}, adding type supervision alone improves DIRL from 84.1 to 87.2 CIDEr. However, this gain remains much smaller than that of DIRL+MEDIC, which reaches 99.8 CIDEr. This result confirms that the performance improvement of MEDIC is not merely due to using change-type labels during training. Instead, the explicit type-specialized expert architecture, together with routing-based expert selection, provides a stronger mechanism for learning change-type-specific reasoning.}

\section{ {Leave-One-Type-Out Evaluation for In-Domain Held-Out Change Types}}
\label{sec:held_out_type}
 {To examine how MEDIC handles in-domain change types that are not observed during training and how it can later incorporate them with additional supervision, we conduct a leave-one-type-out evaluation on CLEVR-DC.}

 {Specifically, CLEVR-DC provides a predefined change-type taxonomy, and we exclude one existing type, such as `Color' or `Add', during training. At test time, samples from the excluded type are evaluated without having trained a dedicated expert for that type.}

 {As shown in Table~\ref{table:leave_one_type_out}, when `Color' is excluded during training, CIDEr on the held-out type drops from 118.8 to 18.4. A similar drop is observed when `Add' is held out, where the score decreases from 86.1 to 13.1. This behavior is expected because no dedicated expert has been optimized for the excluded type. Nevertheless, MEDIC does not completely fail: it can still generate captions by softly routing held-out samples to the most relevant previously learned experts.}

 {We then simulate an incremental update scenario where supervision for the held-out type becomes available after the initial training stage. In this setting, we freeze the existing experts, add a newly initialized expert for the held-out type, and update the expanded router together with the new expert. This requires only 0.2M additional parameters, corresponding to 0.82\% of the 25.12M total parameters. After this lightweight update, performance on the held-out type substantially recovers: `Color' improves from 18.4 to 113.4 and `Add' improves from 13.1 to 86.6, approaching the fully supervised setting trained with all six types. (118.8 and 86.1)} 

 {These results support the modular extensibility of MEDIC. MEDIC substantially recovers performance on the held-out type while avoiding catastrophic degradation on previously learned types. This demonstrates that newly supervised change categories can be incorporated through lightweight expert-level fine-tuning rather than full model retraining. Overall, our leave-one-type-out evaluation shows two complementary strengths: (1) partial robustness to in-domain held-out types through soft expert routing, and (2) efficient extensibility via modular expert addition.}

\begin{table}[t]
    \renewcommand{\tabcolsep}{2.3mm}
    \centering
    \caption{ {Leave-one-type-out results on CLEVR-DC in an in-domain held-out type setting. FT denotes incremental fine-tuning after adding a new expert for the held-out type while preserving the existing experts.}}
    \resizebox{\linewidth}{!}{
        \begin{tabular}{lccccccccc}
        \toprule
        \textbf{Method} & \textbf{\# Train Types} & \textbf{Unseen} & 
        \textbf{Total} & \textbf{Clr} & \textbf{Tex} & \textbf{Add} & 
        \textbf{Drp} & \textbf{Mv} & \textbf{NC} \\
        \midrule
        DIRL & 6 & -- & 84.1 & 108.6 & 76.0 & 76.5 & 81.5 & 38.8 & 41.6 \\
        DIRL+MEDIC & 6 & -- & 99.8 &  {\underline{118.8}} & 88.2 &  {\underline{86.1}} & 88.3 & 54.4 & 82.3 \\
        \midrule
        (1) DIRL+MEDIC($\mathcal{B}$) & 5 & Color & 76.1 & \textbf{18.4} & 85.7 & 87.0 & 84.7 & 39.6 & 56.0 \\
        (2) ($\mathcal{B}$)+FT Color & 6 & -- & 92.0 & \textbf{113.4} & 85.1 & 80.5 & 85.1 & 48.0 & 70.1 \\
        \midrule
        (1) DIRL+MEDIC($\mathcal{B}$) & 5 & Add & 74.5 & 117.5 & 85.9 & \textbf{13.1} & 83.3 & 39.4 & 54.3 \\
        (2) ($\mathcal{B}$)+FT Add & 6 & -- & 93.7 & 114.9 & 85.6 & \textbf{86.6} & 82.8 & 53.2 & 63.6 \\
        \bottomrule
        \end{tabular}}
    \label{table:leave_one_type_out}
\end{table}

\section{Taxonomy-Mismatch Stress Test}
\label{sec:taxonomy-mismatch}
To evaluate robustness under taxonomy mismatch, we conduct a stress test in which the predefined change-type taxonomy is intentionally perturbed. Specifically, the Color type is removed and its samples are reassigned to other categories using an LLM, introducing ambiguous and potentially inconsistent type labels during training. Importantly, this setting does not remove color-change instances themselves. The underlying image pairs and their caption ground-truths still contain color modifications, while only the type labels provided to the model are mismatched.

As shown in Table~\ref{table:taxonomy_mismatch}, this label inconsistency reduces routing accuracy because the router is trained under ambiguous supervision. However, captioning performance degrades only moderately, and the model continues to describe many color changes correctly. This outcome is reasonable since MEDIC still observes color-change examples during training; although their type labels are reassigned, the model can learn from the visual evidence in the input images rather than relying solely on discrete type annotations.

Overall, these results indicate that MEDIC remains functional under taxonomy-level ambiguity. Even when type supervision is imperfect, the expert modules retrieve and integrate visual cues directly from the images, enabling stable change captioning despite mismatched change-type definitions.

\begin{table}[t]
    \renewcommand{\tabcolsep}{3mm}
    \centering
    \caption{Taxonomy-mismatch stress test.}
    \resizebox{0.8\linewidth}{!}{
        \begin{tabular}{l c c c c}
        \toprule
        \textbf{Method} & \textbf{Reclassified} & \textbf{Total} & \textbf{Color} & \textbf{Routing Acc.} \\
        \midrule
        DIRL & - & 84.1 & 108.6 & - \\
        \cdashline{1-5}
        \addlinespace[1.2pt]
        \multirow{2}{*}{DIRL+MEDIC}
        & - & 99.8 & 118.8 & 75.6 \\
        & Color & 87.5 & 115.7 & 60.4 \\
        \bottomrule
        \end{tabular}}
    \label{table:taxonomy_mismatch}
\end{table}

\begin{figure}[t]
    \begin{minipage}[b]{0.6\linewidth}
	\centering
        \centerline{\includegraphics[width=\linewidth]{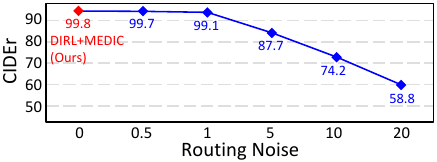}}
        \end{minipage}
    \centering
	\caption{Effect of Routing Accuracy.}
    \label{fig:routing_acc}
\end{figure}

\section{Impact of Routing Accuracy on Caption Quality}
\label{sec:routing acc and caption quality}
To investigate how routing accuracy affects caption quality, we simulate different routing accuracies by adding Gaussian noise $\sigma \in \{0, 0.5, 1, 5, 10, 20\}$. As shown in Figure~\ref{fig:routing_acc}, when routing noise is small ($\sigma \leq 1$), caption performance remains nearly unchanged, indicating that the model is robust to minor routing perturbations. However, as the noise level increases (indicating reduced routing accuracy), the CIDEr score decreases steadily, dropping significantly when the routing decisions become highly unreliable. These results reveal a clear relationship between routing reliability and caption generation quality. Accurate routing enables the model to leverage the appropriate type-specific experts, which in turn leads to more precise change descriptions.




\begin{table}[t]
\renewcommand{\tabcolsep}{2.2mm}
\centering
\caption{ {Additional change-aware evaluation on CLEVR-DC using DIRL as the backbone. Chg/NC Acc. measures whether the generated caption correctly distinguishes changed and unchanged cases. Type Recog. measures whether the described change type matches the ground-truth type. Content Align. is measured by CLIPScore.}}
\label{table:detailed_eval}
\resizebox{0.90\linewidth}{!}{
\begin{tabular}{lccc}
\toprule
\textbf{ {Method}} &
\textbf{ {Chg/NC Acc.}} &
\textbf{ {Type Recog.}} &
\textbf{ {Content Align.}} \\
\midrule
 {DIRL} &
 {83.5} &
 {77.8} &
 {26.5} \\
 {DIRL+MEDIC} &
\textbf{ {92.4}} &
\textbf{ {84.7}} &
\textbf{ {27.2}} \\
\bottomrule
\end{tabular}
}
\end{table}

\section{Ablation of Routing Strategy}
\label{sec:routing_strategy}
To validate the effectiveness of our routing design, we compare the proposed two-stage routing strategy with a simplified single-stage alternative on CLEVR-DC. In the proposed two-stage routing, the model first determines whether a change exists using a change router (Stage-1). If a change is detected, a type router (Stage-2) then predicts the specific change type. This hierarchical structure separates change detection from fine-grained type classification. In contrast, the single-stage variant performs routing in a single step over all six categories, including the no-change class. That is, one router directly predicts among all change types and the no-change label without an explicit decomposition. Under this setting, the single-stage routing achieves a CIDEr score of 90.4, whereas the proposed two-stage routing reaches 99.8. This substantial performance gap confirms that explicitly decomposing change detection and type classification leads to more effective routing and improved caption quality.

\section{ {Additional Evaluation Beyond Caption Similarity Metrics}}
\label{sec:detailed_evaluation}

 {Standard captioning metrics such as BLEU-4~\cite{bleu}, METEOR~\cite{meteor}, ROUGE-L~\cite{rouge}, CIDEr~\cite{cider}, and SPICE~\cite{spice} evaluate similarity between generated captions and reference captions, but they do not directly isolate whether a model correctly recognizes the existence of change, the type of change, or the detailed visual content. To complement these metrics, we conduct additional change-aware evaluations on CLEVR-DC using DIRL as the backbone.}

 {We consider three complementary measures. First, change/no-change accuracy (Chg/NC Acc.) checks whether the generated caption correctly distinguishes changed cases from no-change cases. Second, change-type recognition (Type Recog.) measures whether the generated caption describes the same change type as the ground-truth label. For these two measures, we use a rule-based parser based on no-change expressions and type-specific keywords that commonly appear in CLEVR-DC captions. Third, content alignment (Content Align.) is measured using CLIPScore, which evaluates image-caption alignment and provides a complementary signal for detailed content correctness.}

 {As shown in Table~\ref{table:detailed_eval}, DIRL+MEDIC outperforms DIRL across all three evaluations. In particular, MEDIC improves change/no-change accuracy from 83.5 to 92.4 and change-type recognition from 77.8 to 84.7, indicating that type-specialized experts help the model better identify whether a change occurred and what type of change occurred. MEDIC also improves CLIPScore from 26.5 to 27.2, suggesting improved alignment between the generated caption and visual content. These results support that MEDIC improves not only caption similarity metrics but also change-aware recognition behavior.}

\begin{table}[t]
\renewcommand{\tabcolsep}{2.2mm}
\centering
\caption{ {Comparison with LVLM-based methods on Spot-the-Diff. Zero-shot LVLM results are reported from prior work~\cite{VPG-C}. BLIP2IDC is a fine-tuned LVLM-based model.}}
\label{table:mllm_spot}
\resizebox{0.95\linewidth}{!}{
\begin{tabular}{lccc}
\toprule
\textbf{ {Method}} & 
\textbf{ {Setting}} & 
\textbf{ {METEOR}} & 
\textbf{ {CIDEr}} \\
\midrule
 {BLIP-2~\cite{li2023blip}} & 
 {Zero-shot} & 
 {--} & 
 {17.5} \\
 {InstructBLIP~\cite{dai2023instructblip}} & 
 {Zero-shot} & 
 {--} & 
 {19.7} \\
 {VPG-C-LLaMA2-7B~\cite{VPG-C}} & 
 {Zero-shot} & 
 {--} & 
 {21.0} \\
 {VPG-C-Vicuna-7B~\cite{VPG-C}} & 
 {Zero-shot} & 
 {--} & 
 {20.0} \\
 {VPG-C-Vicuna-13B~\cite{VPG-C}} & 
 {Zero-shot} & 
 {--} & 
 {21.6} \\
\midrule
 {BLIP2IDC~\cite{blip2idc}} & 
 {Fine-tuned LVLM} & 
 {13.5} & 
\textbf{ {51.4}} \\
 {DIRL+MEDIC} & 
 {Task-trained lightweight model} & 
\textbf{ {14.6}} & 
 {45.5} \\
\bottomrule
\end{tabular}
}
\end{table}


\section{ {Comparison with LVLM-Based Methods}}
\label{sec:mllm_comparison}

 {Recent LVLM-based approaches have shown strong potential for image difference captioning by leveraging large-scale vision-language pretraining. To better position MEDIC with respect to these methods, we compare it with both zero-shot LVLMs and a fine-tuned LVLM-based change captioning model.}

 {Table~\ref{table:mllm_spot} summarizes results on Spot-the-Diff. Zero-shot LVLMs such as BLIP-2~\cite{li2023blip}, InstructBLIP~\cite{dai2023instructblip}, and VPG-C~\cite{VPG-C} variants achieve CIDEr scores between 17.5 and 21.6. These results indicate that directly applying general-purpose LVLMs in a zero-shot manner remains challenging for change captioning, where the model must precisely compare two similar images and describe only the visual difference.}

 {When task-specific fine-tuning is introduced, the LVLM-based BLIP2IDC~\cite{blip2idc} achieves a much higher CIDEr score of 51.4, showing the importance of adapting LVLMs to the change captioning task. Compared with this fine-tuned LVLM-based model, DIRL+MEDIC obtains a lower CIDEr score but a higher METEOR score. This suggests that MEDIC remains competitive with fine-tuned LVLM-based approaches while using a lightweight task-trained backbone and a type-aware reasoning module, without relying on large-scale LVLM pretraining.}

\section{Qualitative Results}
\label{sec:qualitative results}

As mentioned in the main paper (Section 4.7) \textbf{``Qualitative Results''},
We provide additional visualizations for each dataset to further illustrate the effects of change-type awareness. We present qualitative examples in Figure~\ref{fig:qualitative_datasets}, \ref{fig:qualitative_datasets2}, \ref{fig:qualitative_datasets3} to demonstrate how the proposed MEDIC, through type-aware modeling, more accurately identifies and localizes changes, leading to more detailed and precise descriptions of the changed regions. Across all cases, MEDIC enhances ability of the model to generate \textit{type-aware} and \textit{semantically accurate} descriptions, particularly in challenging scenarios where previous methods often fail to capture the correct change type.

\section{ {Failure Cases}}
\label{sec:failure cases}

 {While MEDIC generally improves change captioning performance across datasets, it still exhibits several characteristic failure modes. Figure~\ref{fig:failure_cases} presents representative failure cases from CLEVR-Change, Spot-the-Diff, and Image-Editing-Request.}

 {In CLEVR-Change, MEDIC may fail when the visual change is subtle and localized, such as a small object movement under limited appearance variation. In such cases, the changed object can be difficult to align reliably between the before and after images, leading the model to incorrectly predict that the scene remains unchanged.}

 {In Spot-the-Diff, failure often arises in crowded scenes containing multiple entities and potentially multiple simultaneous changes. Under these conditions, the model may generate a partially correct but incomplete caption that describes salient objects in the after image without accurately capturing the underlying change relation, such as movement or disappearance.}

 {In Image-Editing-Request, failures are frequently associated with global editing operations whose semantic boundaries are ambiguous, such as contrast adjustment, color enhancement, or style-related appearance changes. In these cases, MEDIC may produce an edit description that is semantically related to the target instruction but does not precisely match the intended operation.}

 {Overall, these examples suggest that the main remaining challenges for MEDIC include subtle local changes, complex multi-object scenes, and ambiguous global edits. Addressing these limitations more effectively remains an important direction for future work.}

\section{Dynamic Slot Activation Analysis}
\label{sec:dynamic slot activation}

To complement the analysis presented in the main paper (Section 4.6) \textbf{``Dynamic Slot Activation''}, we provide additional visualizations across various change types to further validate the dynamic behavior of our memory-based experts.
Figures~\ref{fig:slot-activation-part1} and \ref{fig:slot-activation-part2} show memory activation patterns for two samples per change type from the CLEVR-DC dataset \cite{clevrdc}. 

Although the samples belong to the same change type, we observe notable differences in the activated memory slots across examples.
In particular, we observe two distinct types of slot activations: \textbf{type-specific slots} (green circles) and \textbf{input-specific slots} (blue circles). Type-specific slots are consistently activated across different input samples that share the same change type, indicating that they capture generalizable patterns associated with that type. In contrast, input-specific slots are selectively activated based on the unique visual context of each input. These input-dependent activations allow the expert to encode fine-grained, input-specific details.

As a result, these findings highlight that the memory-based experts do not rely on static mappings. Instead, they dynamically retrieve relevant information by selecting different memory slots in an input-conditioned manner, demonstrating both flexibility and semantic sensitivity in slot utilization.
It also provides further evidence that memory-based retrieval enables our experts to dynamically specialize without hard-coded priors or slot assignments.


\begin{figure*}[t]
    \centering

    \begin{subfigure}{1\linewidth}
        \centering
        \includegraphics[width=\linewidth]{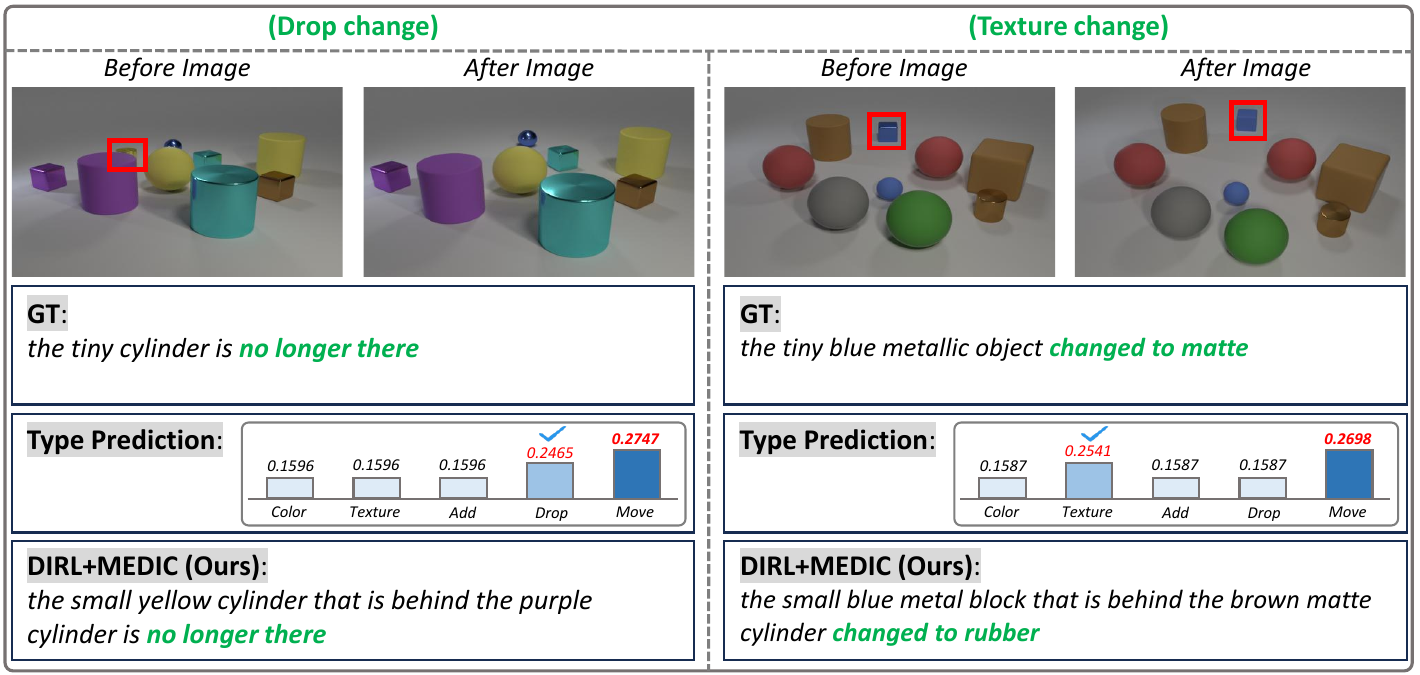}
        \caption{CLEVR-Change dataset.}
    \end{subfigure}
    \vspace{1.5mm}

    \begin{subfigure}{1\linewidth}
        \centering
        \includegraphics[width=\linewidth]{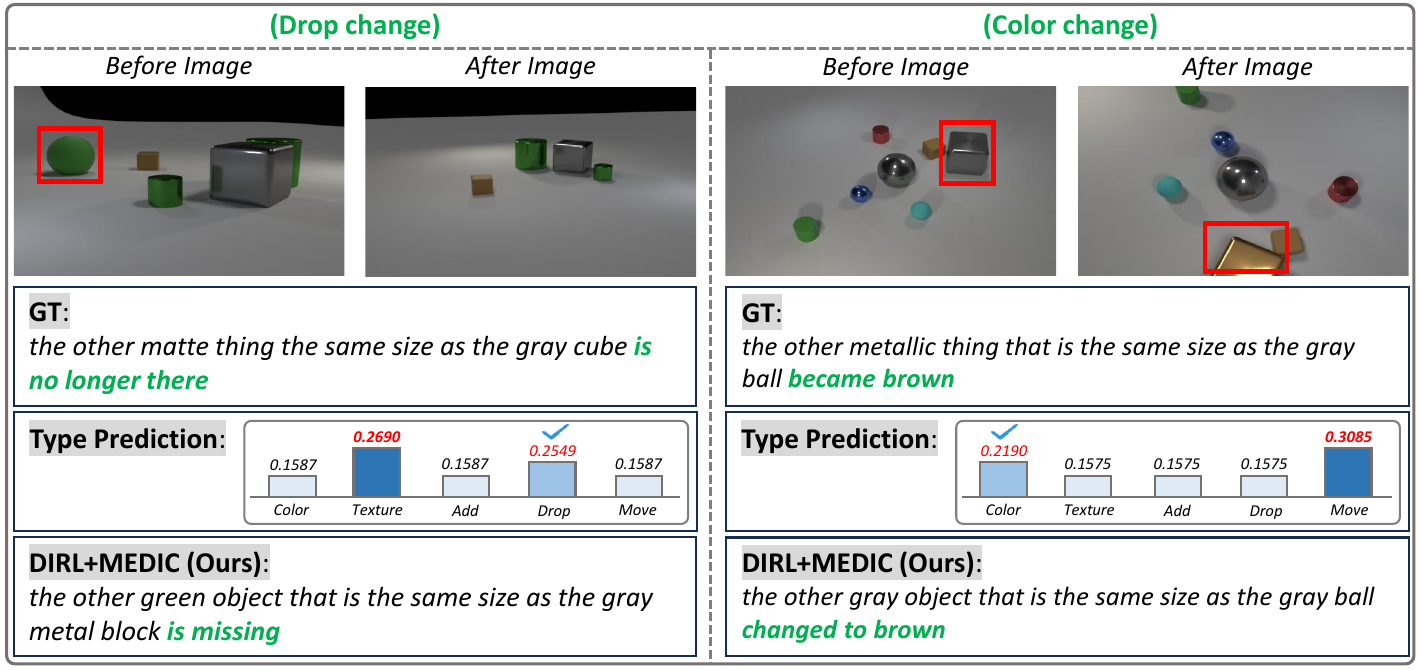}
        \caption{CLEVR-DC dataset.}
    \end{subfigure}
    \vspace{1.5mm}

    \begin{subfigure}{1\linewidth}
        \centering
        \includegraphics[width=\linewidth]{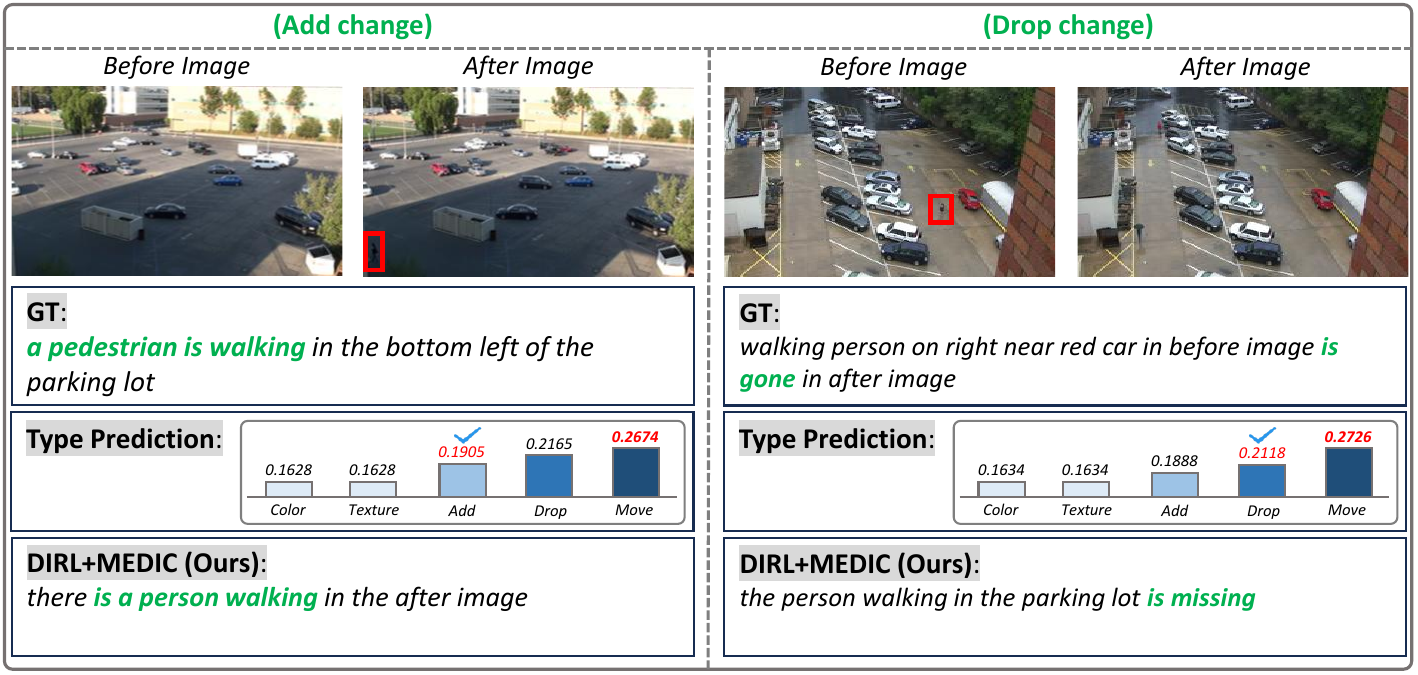}
        \caption{Spot-the-Diff dataset.}
    \end{subfigure}

    \caption{
    Qualitative analysis of cases where the router's top-1 predicted change type differs from the ground-truth type. Across CLEVR-Change, CLEVR-DC, and Spot-the-Diff, our model (DIRL+MEDIC) still produces accurate change captions in these challenging cases. This illustrates how the weighted expert aggregation effectively mitigates the impact of non-GT top-1 predictions by integrating complementary cues from all change-type experts.
    }
    \label{fig:router_analysis}
\end{figure*}

\begin{table*}[!t]
    \centering
    \caption{Prompt used for LLM-based generation of change-type labels from ground-truth captions for single-change scenarios in the Spot-the-Diff dataset.}
    \vspace{-0.4cm}
    \begin{mybox2}
    
\textbf{Prompt:}\\
You are given a JSON object where each key is an image ID, and each value is a list of ground truth (GT) captions describing the change between two images associated with that ID. \\

Your task is to classify each image into exactly one of the following six change types: ``\textbf{color}", ``\textbf{texture}", ``\textbf{add}", ``\textbf{drop}", ``\textbf{move}", ``\textbf{no change}". \\
  
Return ONLY the type name (\textit{e.g.,} ``move", ``add", etc.) for the given set of captions. \\

\textbf{[Classification Rules \& Examples]}

(color) \\
- Only color changed. \\ 
- \textbf{Examples}: ``the other large red object that is the same shape as the brown metal object became blue", ``the other ball that is the same size as the cylinder became yellow", ``the other small purple object the same shape as the tiny shiny object became green" \\

(texture) \\
- Material changes (rubber, shiny, matte, metallic). \\
- \textbf{Examples}: ``the other large cylinder the same color as the large matte cylinder turned rubber", ``the other big object the same shape as the tiny yellow rubber thing became shiny", ``the other red sphere the same material as the tiny cyan object became matte" \\

(add) \\
- Something new appears in the after image. \\
- \textbf{Examples}: ``the object behind the green shiny cylinder and right of the large block in the after image has been added", ``the large purple thing has been added", ``the other gray object that is the same size as the cyan object has been newly placed" \\

(drop) \\
- Something from the before image is missing in the after image.\\
- \textbf{Examples}: ``the red cylinder is missing",``the other sphere that is the same size as the blue cube is missing", ``the brown shiny ball is no longer there" \\

(move) \\
- Object's position changed. \\
- \textbf{Examples}: ``the small brown object is in a different location", ``the other shiny ball that is the same size as the blue object is in a different location", ``the other metallic thing the same size as the red cylinder moved"

    \end{mybox2}
     \label{prompt_single_std}
\end{table*}

\begin{table*}[!t]
    \centering
    \begin{mybox2}
    
(no change)\\
- Only assign if all captions explicitly state no difference.\\
- \textbf{Examples}: ``the scene remains the same", ``there is no change", ``nothing was modified"\\

Make sure to think carefully. \\
Just because the word ``color” appears doesn’t mean it’s the color type. The change must actually involve a color transformation to be classified as ``color,” and the same applies to the other types.\\
Don’t rely on keywords alone, understand the context.\\

Now, given the following GT Captions for a single image, respond with one and only one of the following types:\\
``color", ``texture", ``add", ``drop", ``move", or ``no change".\\

    \end{mybox2}
\end{table*}

\begin{table*}[!t]
    \centering
    \caption{Prompt used for LLM-based generation of change-type labels from ground-truth captions for single-change scenarios in the Image-Editing-Request dataset.}
    \vspace{-0.4cm}
    \begin{mybox2}
    
\textbf{Prompt:}\\
You are given a JSON object where each key is an image ID, and each value is a list of ground truth (GT) captions describing the change between a pair of images. \\

Your task is to read the caption and determine the main type of visual change described in it. \\

Focus on understanding the semantic change operation expressed in the caption. \\

Examples of possible change operations include introducing a new element, removing something, replacing one object with another, modifying the background scene, adjusting lighting properties, applying visual effects, or changing the spatial framing of the image. \\

\textbf{Instructions:} \\

1. Carefully read the caption.

2. Infer the \textbf{primary visual editing operation} it describes.

3. Categorize the change at a \textbf{coarse semantic level}, focusing only on the \textbf{main type of change}.

4. \textbf{Do not create overly specific or fine-grained categories}. The goal is to capture the \textbf{general type of change}, not detailed subtypes.

5. If multiple edits are mentioned, choose the \textbf{most dominant or central change}.

6. Output \textbf{only the change category label}. \\

Do not output explanations.

    \end{mybox2}
     \label{prompt_single_ier}
\end{table*}

\begin{table*}[!t]
    \centering
    \caption{Prompt used for LLM-based generation of change-type labels from ground-truth captions in multi-change scenarios.}
    \vspace{-0.3cm}
    \begin{mybox2}

\textbf{Prompt:}\\
You are given a JSON object where each key is an image ID, and each value is a list of ground truth (GT) captions describing changes between a pair of images. \\

Your task is to classify each image into one or more of the following change types: ``\textbf{color}", ``\textbf{texture}", ``\textbf{add}", ``\textbf{drop}", ``\textbf{move}", ``\textbf{no change}". \\

Return ALL applicable types for the given set of captions.  \\
Output MUST be a JSON dictionary with:\\

- key: image ID \\
- value: a list of ALL correct change types (multi-label allowed) \\
- order of types does not matter\\
- do NOT generate any types that are not supported\\
- do NOT output any explanations\\
- do NOT invent differences not mentioned\\

If a caption states multiple different changes, include ALL relevant types. \\

\textbf{[Examples]}

Captions for ``000235.png":
[``the person in the blue shirt is no longer present. the dark grey SUV is gone in the photo on the right"]

\textbf{Output}:
\{ \\
 ``000235.png": [``drop"] \\
\} \\

Captions for ``000103.png":
[
``the blue car has moved. a person appeared in the second picture"
] 

\textbf{Output}: 
\{ \\
``000103.png": [``move", ``add"]  \\
\} \\

Captions for ``000214.png":
[
``people moved from corner crossing street. white car on left side. traffic light red instead of green"
]

\textbf{Output}:
\{ \\
``000214.png": [``move", ``add", ``color"] \\
\} \\

Now, based on the following input JSON, return only the final JSON result:\\

    \end{mybox2}
     \label{prompt_multi}
\end{table*}

\begin{figure*}[t]
    \centering

    \begin{subfigure}{1\linewidth}
        \centering
        \includegraphics[width=\linewidth]{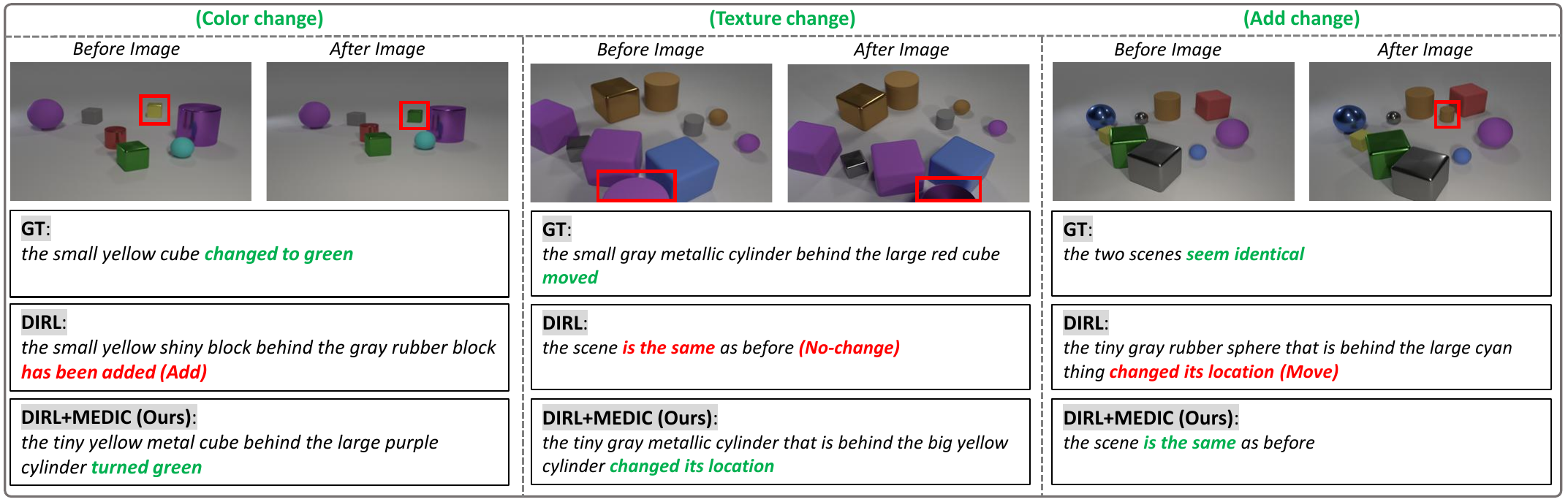}
    \end{subfigure}
    \begin{subfigure}{1\linewidth}
        \centering
        \includegraphics[width=\linewidth]{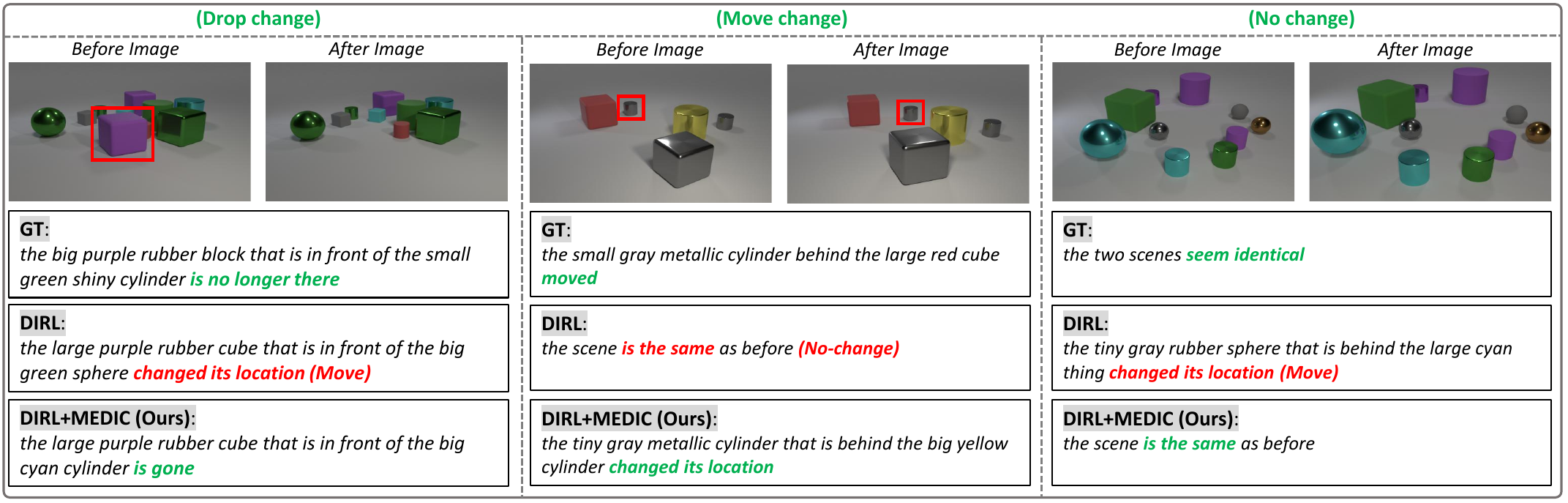}
        \caption{CLEVR-Change dataset.}
    \end{subfigure}

    \vspace{0.4em}

    \begin{subfigure}{1\linewidth}
        \centering
        \includegraphics[width=\linewidth]{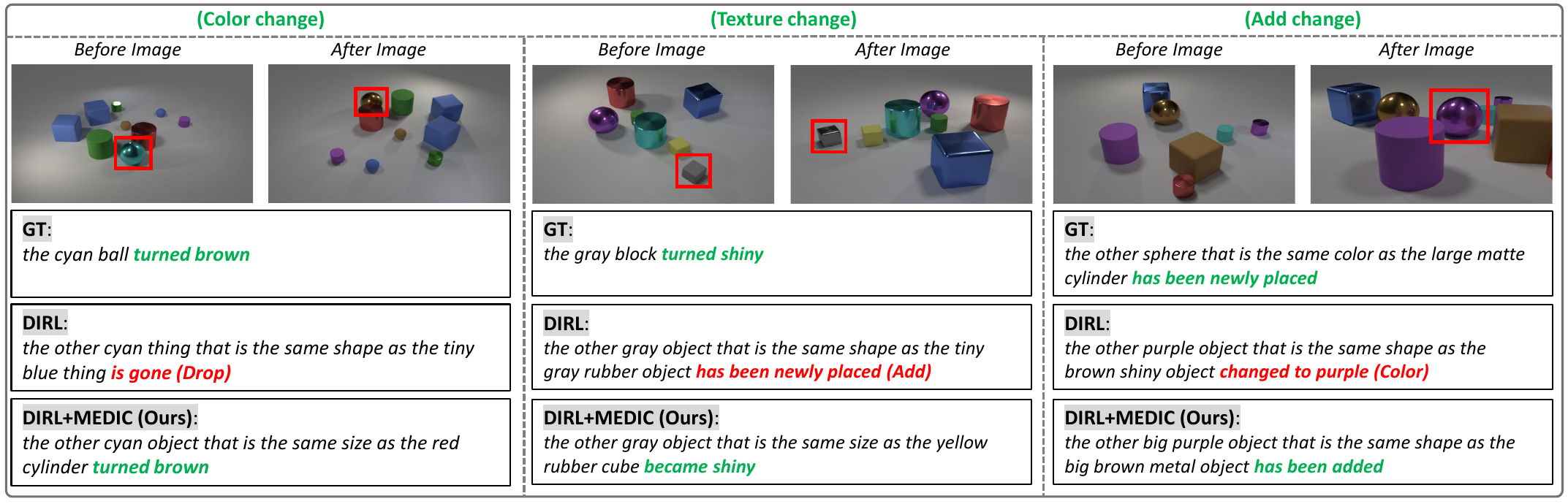}
    \end{subfigure}
    \begin{subfigure}{1\linewidth}
        \centering
        \includegraphics[width=\linewidth]{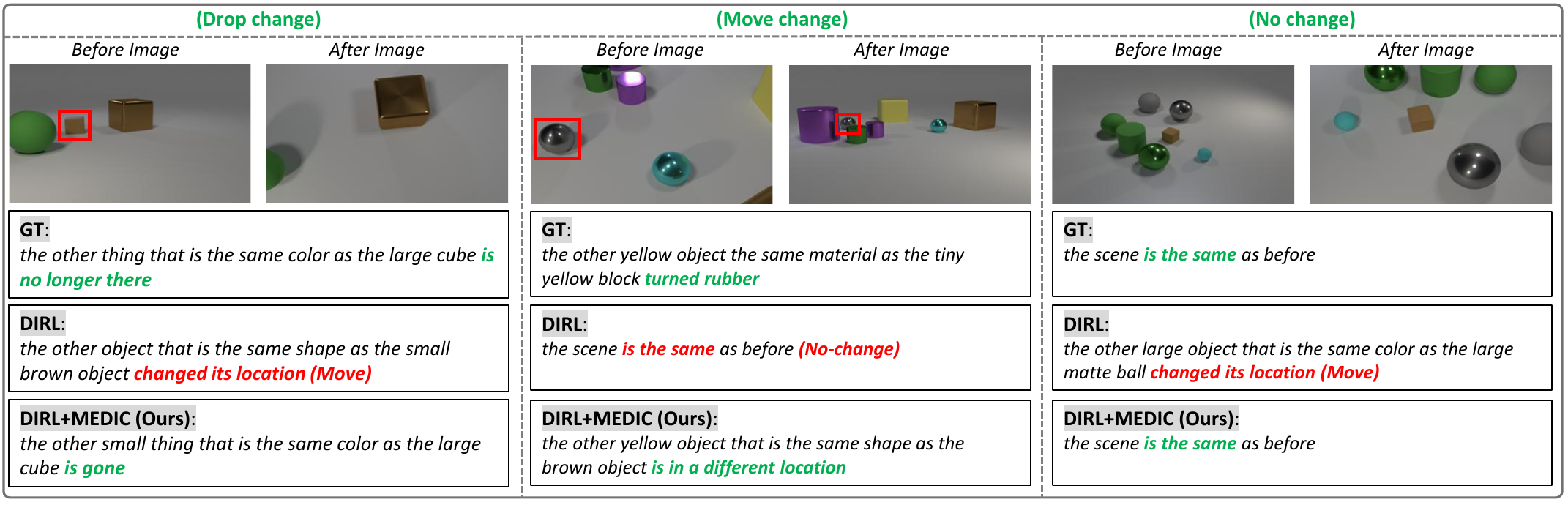}
        \caption{CLEVR-DC dataset.}
    \end{subfigure}

    \vspace{-0.6em}
    \caption{\textbf{
    Qualitative examples comparing GT, DIRL, and DIRL+MEDIC (Ours) across datasets (Part 1)}.
    Our model generates more accurate and type-consistent captions by explicitly modeling change types.
    }
    \label{fig:qualitative_datasets}
\end{figure*}

\begin{figure*}[t]
    \centering

    \begin{subfigure}{1\linewidth}
        \centering
        \includegraphics[width=\linewidth]{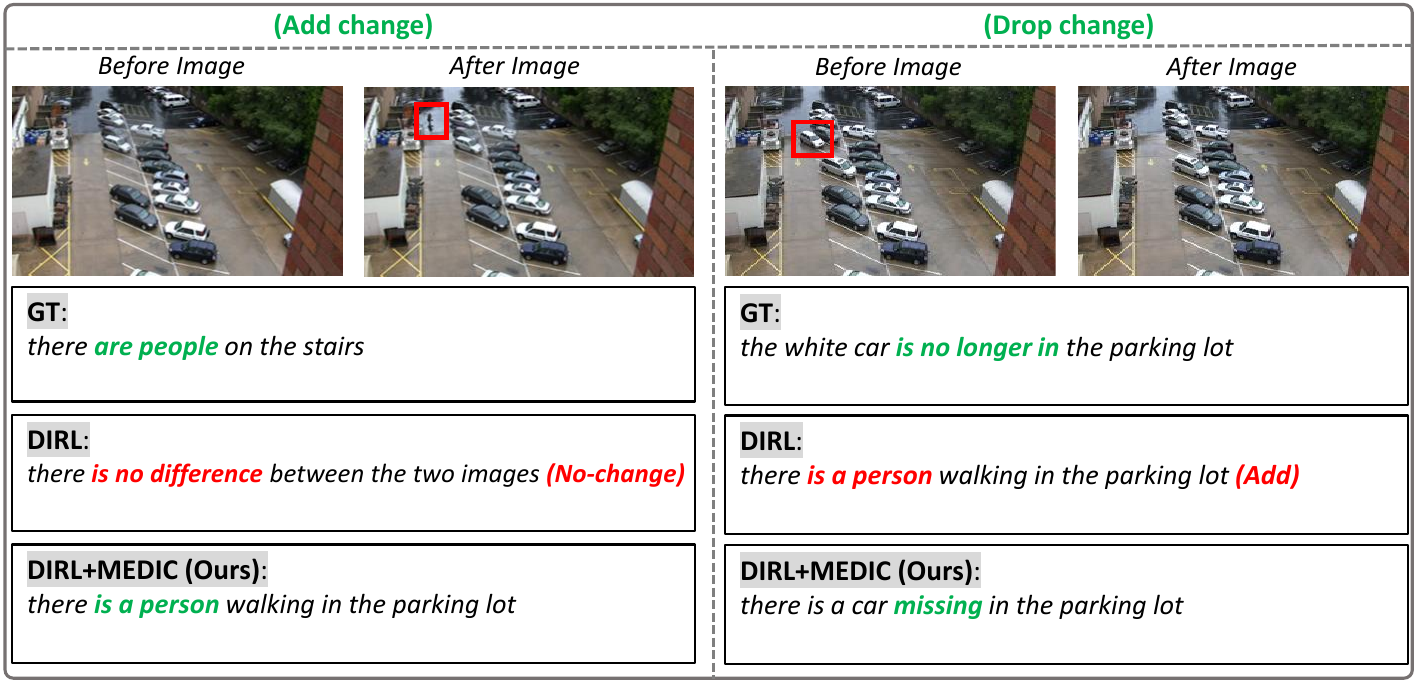}
    \end{subfigure}
    \begin{subfigure}{1\linewidth}
        \centering
        \includegraphics[width=\linewidth]{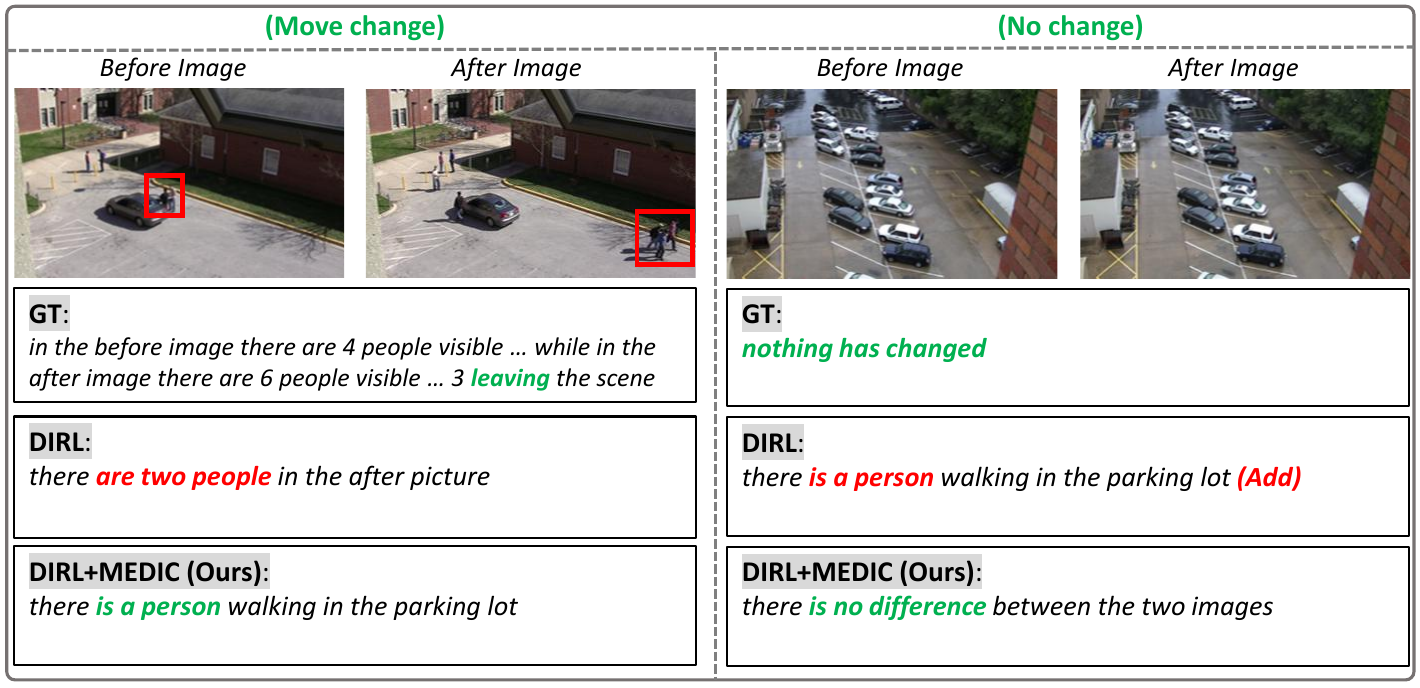}

        \setcounter{subfigure}{2}
        \caption{Spot-the-Diff dataset.}
    \end{subfigure}

    \caption{\textbf{
    Qualitative examples comparing GT, DIRL, and DIRL+MEDIC (Ours) across datasets (Part 2)}.
    Our model generates more accurate and type-consistent captions by explicitly modeling change types.
    }
    \label{fig:qualitative_datasets2}
\end{figure*}

\begin{figure*}[t]
    \centering

    \begin{subfigure}{1\linewidth}
        \centering
        \includegraphics[width=\linewidth]{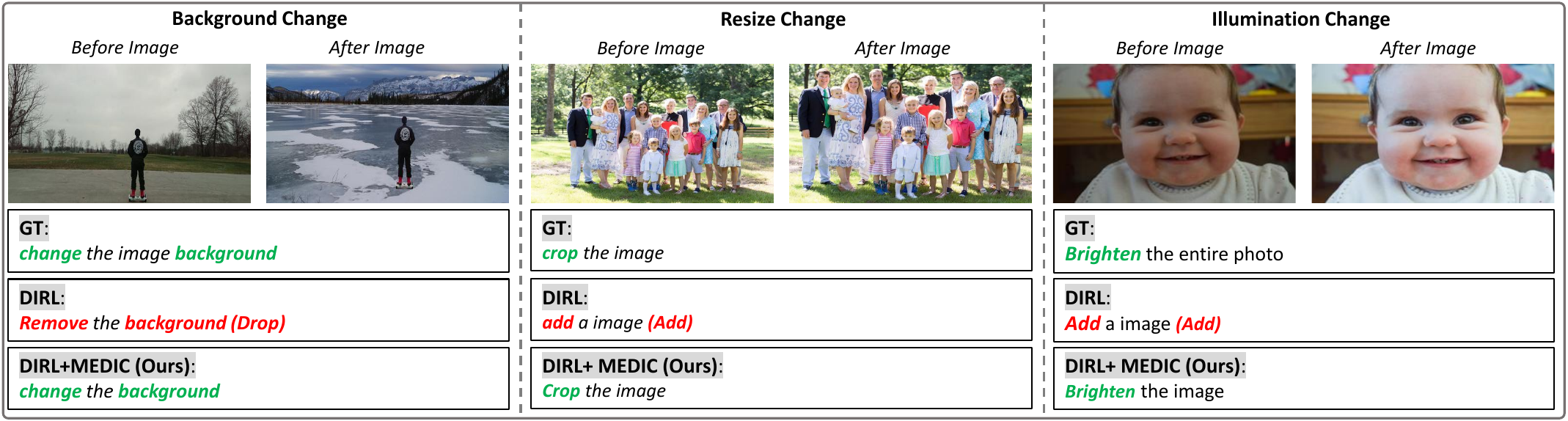}

        \setcounter{subfigure}{3}
        \caption{Image Editing Request dataset.}
    \end{subfigure}

    \caption{\textbf{
    Qualitative examples comparing GT, DIRL, and DIRL+MEDIC (Ours) across datasets (Part 3)}.
    Our model generates more accurate and type-consistent captions by explicitly modeling change types.
    }
    \label{fig:qualitative_datasets3}
\end{figure*}
\begin{figure*}[t]
\centering
\includegraphics[width=\textwidth]{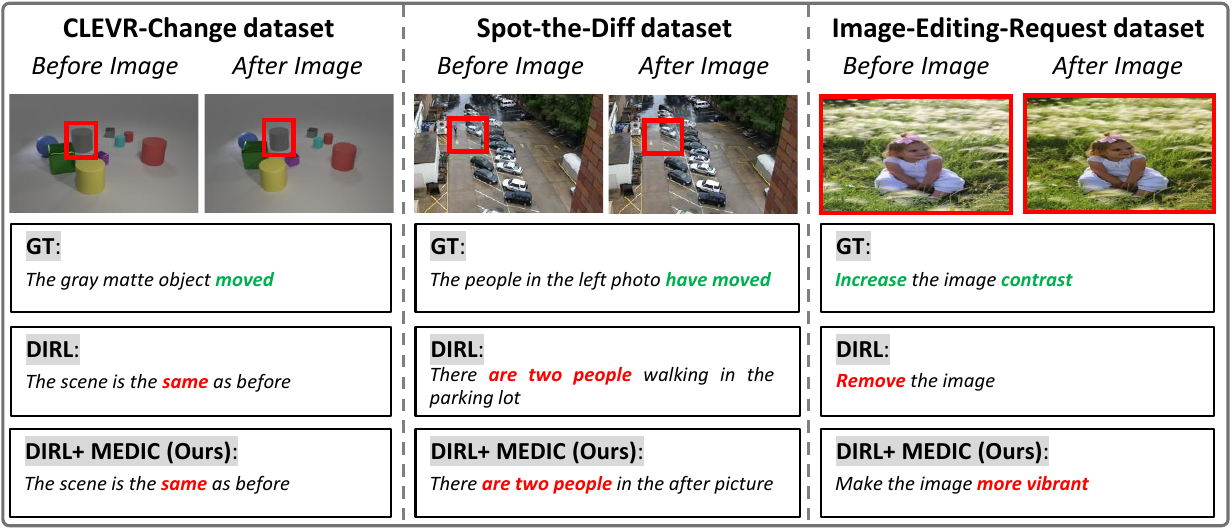}
\caption{ {Representative failure cases of MEDIC on CLEVR-Change, Spot-the-Diff, and Image-Editing-Request. From left to right, the examples illustrate failures under subtle local changes, crowded multi-object scenes, and ambiguous global edit instructions.}}
\label{fig:failure_cases}
\end{figure*}

\begin{figure*}[t]
    \centering
    \includegraphics[width=1\linewidth]{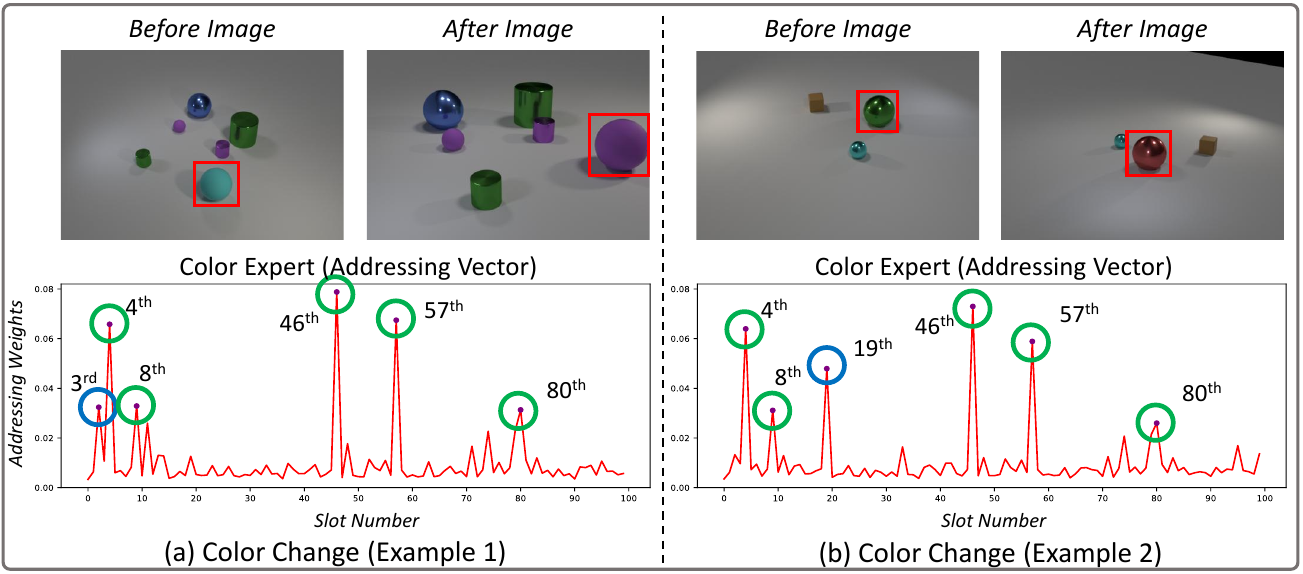}
    \includegraphics[width=1\linewidth]{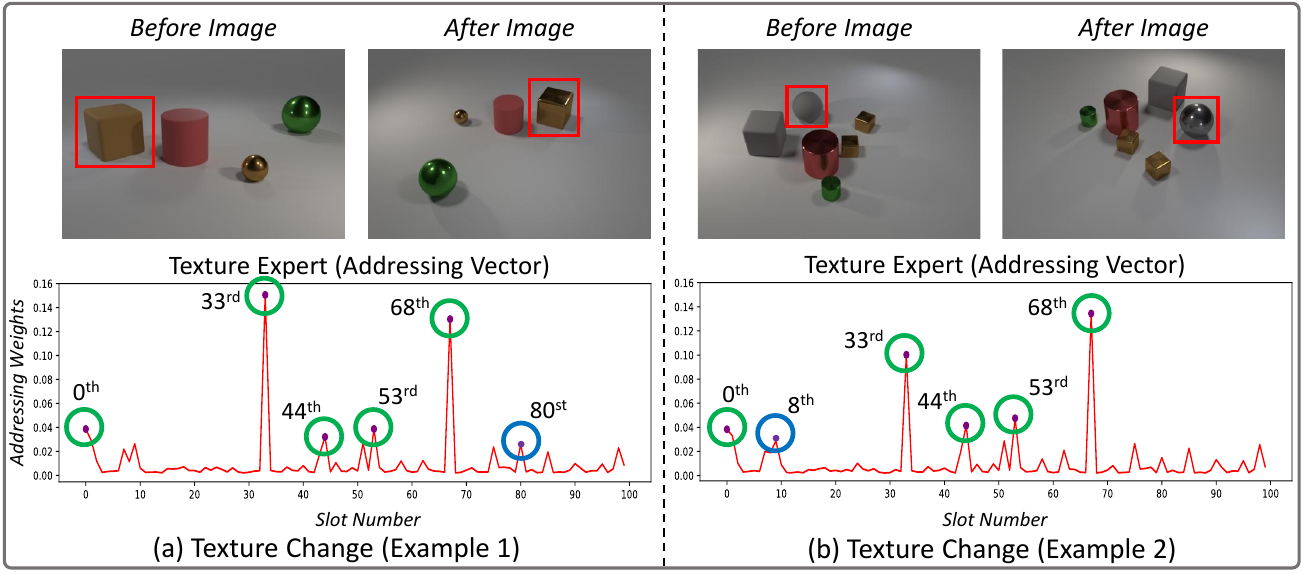}
    \includegraphics[width=1\linewidth]{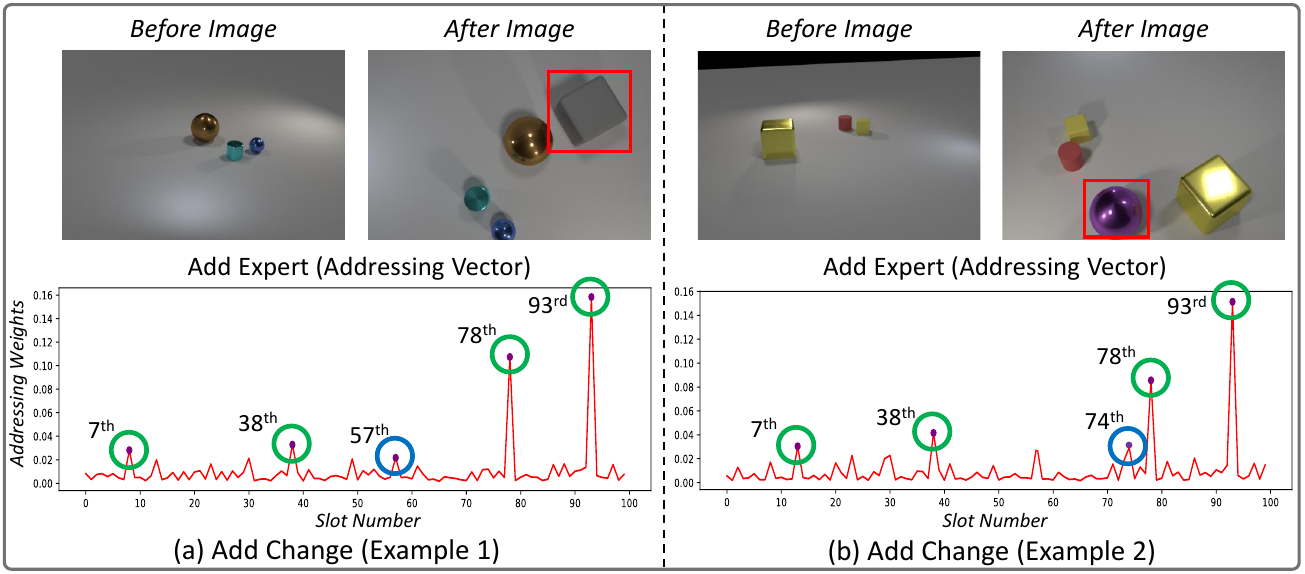}
    \caption{
    \textbf{Dynamic slot activation examples on CLEVR-DC dataset using DIRL+MEDIC (Ours) (Part 1).}  
    For each sample, we show the input image pair (top) and the addressing vector from the selected expert (bottom).  
    \textcolor{green}{Green-circled} peaks represent \textit{type-specific} memory slots that consistently appear across different samples of the same change type, while  {blue-circled} peaks reflect \textit{input-specific} slots adapted to the particular input.
    }
    \label{fig:slot-activation-part1}
\end{figure*}

\begin{figure*}[p]
    \centering
    \includegraphics[width=1\linewidth]{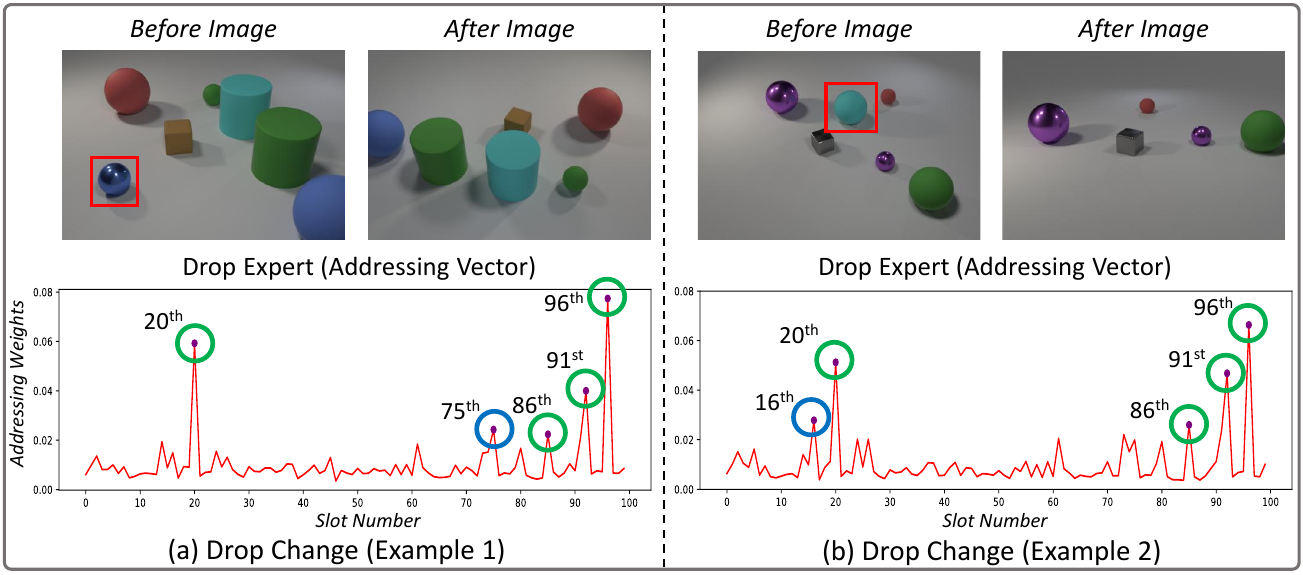}
    \includegraphics[width=1\linewidth]{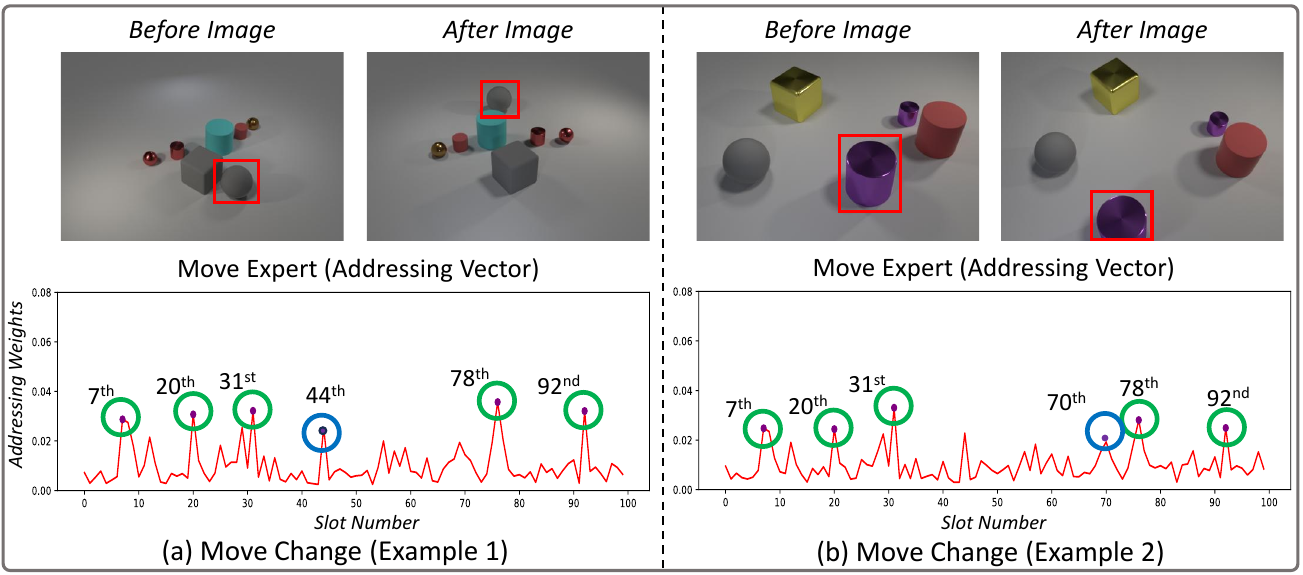}
    \includegraphics[width=1\linewidth]{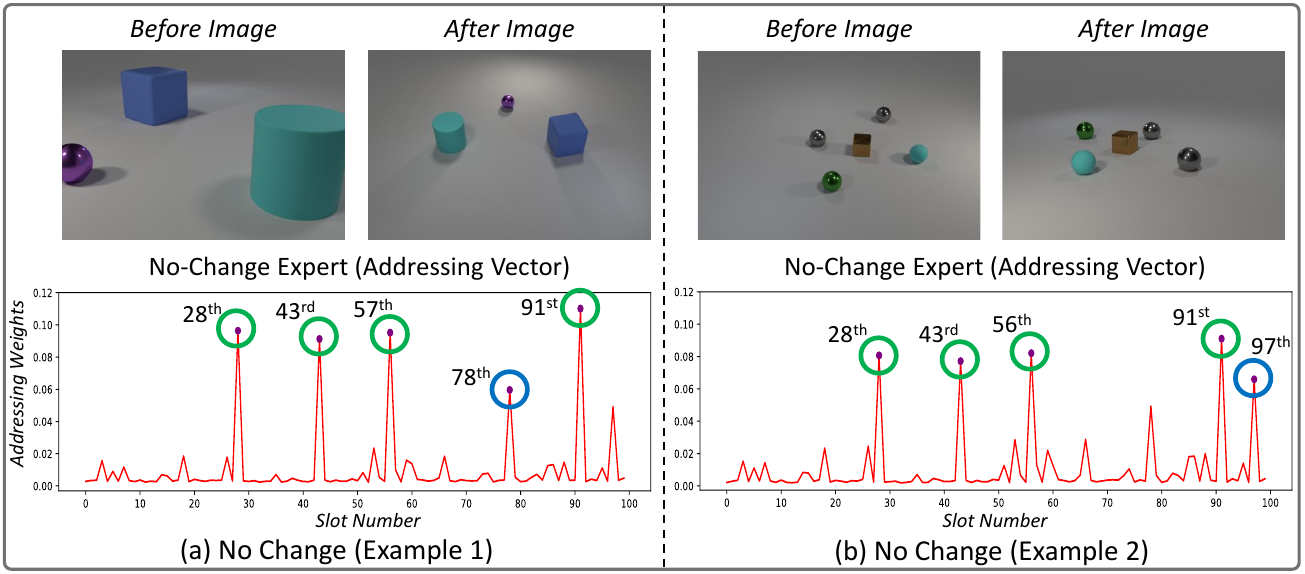}
    \caption{
    \textbf{Dynamic slot activation examples on CLEVR-DC dataset using DIRL+MEDIC (Ours) (Part 2).}  
    For each sample, we show the input image pair (top) and the addressing vector from the selected expert (bottom).  
    \textcolor{green}{Green-circled} peaks represent \textit{type-specific} memory slots that consistently appear across different samples of the same change type, while  {blue-circled} peaks reflect \textit{input-specific} slots adapted to the particular input.
    }
    \label{fig:slot-activation-part2}
\end{figure*}

\clearpage
\newpage

\end{document}